%% file: main.tex
\documentclass[nonacm,sigconf,natbib=false]{acmart}
\AtBeginDocument{%
  }

\setcopyright{acmlicensed}
\copyrightyear{2018}
\acmYear{2018}
\acmDOI{XXXXXXX.XXXXXXX}
\acmConference[MMSys]{ACM Multimedia Systems Conference}{June 03--05,
  2018}{Woodstock, NY}
\acmISBN{978-1-4503-XXXX-X/2018/06}

\acmSubmissionID{195}

\RequirePackage[
  datamodel=acmdatamodel,
  style=acmnumeric,
  ]{biblatex}

\usepackage{graphicx} 
\usepackage{caption} 
\usepackage{amsmath}

\usepackage{booktabs}
\usepackage{subcaption}
\usepackage[table]{xcolor}   
\usepackage{multirow}
\usepackage{makecell}

\usepackage[per-mode=symbol]{siunitx}
\usepackage{enumitem}
\usepackage{gensymb}
\usepackage{xspace}
\setlist[itemize]{leftmargin=*}

\usepackage[capitalise,noabbrev]{cleveref}

\input{preamble}
\makeatletter
\DeclareRobustCommand\onedot{\futurelet\@let@token\@onedot}
\def\@onedot{\ifx\@let@token.\else.\null\fi\xspace}
\def\eg{\emph{e.g}\onedot} 
\def\ie{\emph{i.e}\onedot}

\makeatother

\DeclareSIUnit\px{px}
\newcommand{\name}{LiftGS}

\title{Lifting 3D Gaussians: A Wavelet Multiresolution Representation for Scalable Gaussian Splatting Streaming}

\author{Yuang Shi}
\authornote{Also with IPAL, IRL2955, Singapore.}
\affiliation{%
  \department{School of Computing}
  \institution{National University of Singapore}
  \country{Singapore}}
\additionalaffiliation{%
  \department{IRIT}
  \institution{Université de Toulouse}
  \country{France}}
\email{yuangshi@u.nus.edu}

\author{Géraldine Morin}
\authornotemark[1]
\affiliation{%
  \department{Institut de Recherche en Informatique de Toulouse (IRIT)}
  \institution{Université de Toulouse}
  \city{Toulouse}
  \country{France}}
\email{geraldine.morin@toulouse-inp.fr}

\author{Simone Gasparini}
\authornotemark[1]
\affiliation{%
  \department{Institut de Recherche en Informatique de Toulouse (IRIT)}
  \institution{Université de Toulouse}
  \city{Toulouse}
  \country{France}}
\email{simone.gasparini@toulouse-inp.fr}

\author{Wei Tsang Ooi}
\authornotemark[1]
\affiliation{%
  \department{School of Computing}
  \institution{National University of Singapore}
  \country{Singapore}}
\email{dcsooiwt@nus.edu.sg}

\renewcommand{\shortauthors}{Shi et al.}
\renewcommand{\shorttitle}{Lifting 3D Gaussians: A Wavelet Multiresolution Representation for Scalable Gaussian Splatting Streaming}

\begin{document}

\input{sec/0-abstract}



\maketitle


\input{sec/1-introduction}

\input{sec/2-related_work}

\input{sec/3-methodology}

\input{sec/4-experiment}

\input{sec/5-conclusion}


\printbibliography

\end{document}

%% file: preamble.tex
\usepackage{subcaption}
\usepackage[per-mode=symbol]{siunitx}
\usepackage{xspace}
\usepackage[table,xcdraw]{xcolor}
\usepackage{enumitem}
\usepackage[normalem]{ulem}
\usepackage{placeins}
\setlist[itemize]{leftmargin=*}
\usepackage{gensymb}
\usepackage[capitalise,noabbrev]{cleveref}
\usepackage{booktabs}
\usepackage{multirow}
\usepackage{colortbl}
\usepackage{makecell}

\makeatletter
\DeclareRobustCommand\onedot{\futurelet\@let@token\@onedot}
\def\@onedot{\ifx\@let@token.\else.\null\fi\xspace}
\DeclareRobustCommand\nodot{\futurelet\@let@token\@nodot}
\def\@nodot{\ifx\@let@token.\else~\null\fi\xspace}

\def\eg{\emph{e.g}\onedot}

\def\ie{\emph{i.e}\onedot}

\DeclareSIUnit\px{px}
\newcommand{\add}[1]{#1}
\newcommand{\better}[1]{\textbf{#1}}

%% file: sec/0-abstract.tex
\begin{abstract}

3D Gaussian Splatting (3DGS) represents a scene as an unstructured set of primitives optimized at a single scale, offering no native support for progressive delivery to heterogeneous devices. Existing scalable approaches organize these primitives into independent quality layers or spatial hierarchies. However, maintaining separate splat representations across scales introduces redundancy, while independent layers have limited ability to correct coarse geometry as finer detail arrives.
We present {\name}, a multiresolution representation that organizes a scene into a coarse Gaussian set and a tree of refinements. Each refinement replaces one coarse Gaussian with two finer children whose attributes are learned relative to their parent. Inspired by wavelet lifting, we encode these changes using a shared detail vector and per-attribute factors that allow the two children to differ asymmetrically from the parent. A client first receives the coarse set, then reconstructs and substitutes children as refinement data arrives. This construction reuses previously transmitted attributes to encode finer detail and removes replaced parents from the active rendering set, reducing both refinement payload and rendering memory.
Extensive experiments on four benchmark datasets show {\name} cuts splat redundancy from over 65\% to under 25\%, reduces transmission payload and GPU memory by up to 2.4$\times$ and 5.5$\times$ over state-of-the-art scalable-3DGS baselines, achieves smooth quality transitions \add{in simulated progressive delivery}, and compresses a further 8.7$\times$ with standard coding at negligible quality loss. \add{Project website: \url{https://yuang-ian.github.io/liftgs/}}.
\end{abstract}

%% file: sec/1-introduction.tex
\section{Introduction} \label{sec:introduction}

3D Gaussian Splatting (3DGS)~\cite{kerbl20233d} enables photorealistic, real-time rendering of captured scenes, making it attractive for immersive media delivery. 
Delivering these scenes to heterogeneous clients introduces challenges: the representation must accommodate changing bandwidth and client resources while users navigate in six degrees of freedom (6DoF)~\cite{viola2023425}. To maintain user experience under limited bandwidth, a practical solution is sending a useful coarse scene first to the client and improve it as more data is delivered.

It imposes three requirements for scalable delivery~\cite{shi20253dds}: i) \textit{quality scalability}, so one representation serves multiple quality levels through incremental transmission; ii) \textit{rendering scalability}, so any received prefix produces a valid scene at cost proportional to data received; and iii) \textit{incremental adaptation}, so continuous responses can be adopted to changing consumption and resource constraints.

However, standard 3DGS provides an unstructured set of splats optimized at a single target resolution, failing to meet the requirements above. It does not specify which data to send first, how subsequent data should refine the received scene, or how the client's rendering state should evolve. These representation choices matter to both sides of a streaming pipeline: the server must supply useful incremental updates, and the client must turn those updates into a scene that fits its rendering budget. In summary, 3DGS lacks a \textit{multiresolution basis}, and the question is how to build one on an unstructured, adaptively-densified Gaussian set.

\begin{figure}[t]%
    \centering

    \begin{subfigure}{.85\linewidth}
        \centering
        \includegraphics[width=0.15\linewidth]{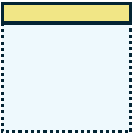}\hfill
        \includegraphics[width=0.15\linewidth]{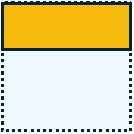}\hfill
        \includegraphics[width=0.15\linewidth]{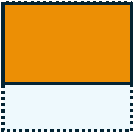}\hfill
        \includegraphics[width=0.15\linewidth]{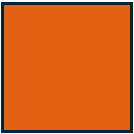}
        \\[2pt] 
        \makebox[0.15\linewidth][c]{$L_0$}\hfill
        \makebox[0.15\linewidth][c]{$L_1$}\hfill
        \makebox[0.15\linewidth][c]{$L_2$}\hfill
        \makebox[0.15\linewidth][c]{$L_3$}
        \caption{Spatial pyramid~\cite{kerbl2024hierarchical,ren2024octree,lu2024scaffold}.}
        \label{subfig:space_hierarchy}
    \end{subfigure}

    \begin{subfigure}{.85\linewidth}
        \centering
        \includegraphics[width=0.15\linewidth]{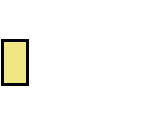}\hfill
        \includegraphics[width=0.15\linewidth]{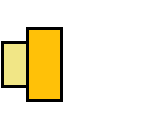}\hfill
        \includegraphics[width=0.15\linewidth]{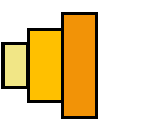}\hfill
        \includegraphics[width=0.15\linewidth]{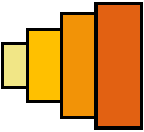}
        \\[2pt]
        \makebox[0.15\linewidth][c]{$L_0$}\hfill
        \makebox[0.15\linewidth][c]{$L_1$}\hfill
        \makebox[0.15\linewidth][c]{$L_2$}\hfill
        \makebox[0.15\linewidth][c]{$L_3$}
        \caption{Independent per-quality layers~\cite{shi2025lapis,tsai2025l3gs,sun2025lts,milef2025learning}.}
        \label{subfig:discrete_hierarchy}
    \end{subfigure}

    \begin{subfigure}{.85\linewidth}
        \centering
        \includegraphics[width=0.95\linewidth]{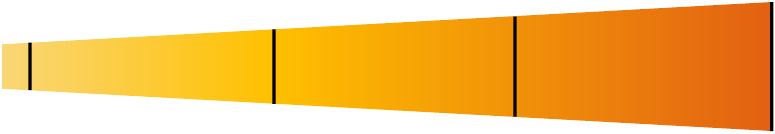}
        \\[2pt]
        \makebox[0.15\linewidth][c]{$L_0$}\hfill
        \makebox[0.15\linewidth][c]{$L_1$}\hfill
        \makebox[0.15\linewidth][c]{$L_2$}\hfill
        \makebox[0.15\linewidth][c]{$L_3$}
        \caption{Our wavelet decomposition.}
        \label{subfig:continuous_hierarchy}
    \end{subfigure}
    \caption{\textbf{Three ways to organize multiscale 3DGS.} \textbf{(a)}~Spatial pyramid: coarse levels merge fine splats bottom-up; each level is a self-contained, redundant summary used for render-time selection. \textbf{(b)}~Independent layers: each quality level adds a separate splat set with no parameter-level relationship, causing error accumulation and redundancy. \textbf{(c)}~Our decomposition: each level is a wavelet refinement of the previous one, generated top-down, critically sampled, and structurally dependent, so quality scales smoothly and proportionally with transmitted data.}%
    \label{fig:hierarchy_illus}
\end{figure}

Existing approaches address different parts of this problem. Spatial hierarchies~\cite{kerbl2024hierarchical,ren2024octree,lu2024scaffold} organize scene content for level-of-detail selection at render time (\cref{subfig:space_hierarchy}). For example, Hierarchical 3DGS~\cite{kerbl2024hierarchical} merges fine splats into coarse summaries. Such hierarchies support efficient rendering, but the full hierarchy must reside on the client, which occupies unnecessary storage resources. Specifically, because each level is a self-contained summary, the hierarchy is an over-complete pyramid (reported over 65\% redundancy~\cite{kerbl2024hierarchical}). Also, a hierarchy alone does not specify a compact incremental transmission format.

Independent-layer methods target progressive delivery directly~\cite{shi2025lapis,tsai2025l3gs,sun2025lts,milef2025learning}. They send a base layer followed by enhancement layers, each containing an additional splat set (\cref{subfig:discrete_hierarchy}), making complete layer prefixes renderable, thus avoiding storing or downloading a full model. 
However, each layer is a fresh splat set, and each is trained to overfit its resolution atop frozen lower layers. Because layers share no parameter-level relationship, errors in coarse layers cannot be corrected, and finer layers must compensate with additional splats or change the opacities of error splats. 
In our evaluation, \SI{65}{\percent} of LapisGS~\cite{shi2025lapis} splats are ``ghost'' splats that are transparent at the finest level, not relevant for rendering yet still contribute to its stored and materialized representation. Quality improvements during partial layer reception are also uneven (\cref{sec:experiment:layer_analysis}).

In this paper, we present {\name}, a multiresolution representation that expresses quality updates as \textit{refinements of already received splats}. We build on adaptive densification in 3DGS~\cite{kerbl20233d}, which uses positional gradients during training to identify poorly reconstructed regions and adds primitives by cloning small splats or splitting large ones. In {\name}, selected coarse splats are split into finer children whose attributes are encoded through a learned parent--child parameterization, which turns densification into a sequence of reconstructible updates. In a streaming scenario, the server sends a coarse splat set, then incrementally transmits the parameters needed to reconstruct finer children from their parents. The client applies each available split by replacing its parent, progressively increasing detail while removing superseded coarse splats from that set. This separates cumulative transmission from active rendering storage: earlier refinements remain part of the encoded scene, while only the current leaf set is needed by the rasterizer.

The refinement design draws on wavelet lifting~\cite{sweldens1996lifting}. In signal and image processing, they define scalable coding solved by wavelet transform. JPEG2000 delivers resolution and quality scalability from one embedded codestream because a wavelet decomposition is multiresolution and critically sampled, ordering a coarse approximation and sparse details with no redundancy~\cite{taubman2002jpeg,shapiro1993embedded,said1996new}. We apply this principle directly to Gaussian attributes.

A parent Gaussian $\boldsymbol{P}$ is a coarse approximation, and it refines into two children $\boldsymbol{C_1}$ and $\boldsymbol{C_2}$ through a learned detail coefficient $\boldsymbol{\psi}$,
$
    \boldsymbol{C_1} = \boldsymbol{P} + \boldsymbol{\psi}, \boldsymbol{C_2} = \boldsymbol{P} - \boldsymbol{\alpha} \odot \boldsymbol{\psi},
$
where $\boldsymbol{\alpha}$ contains five per-attribute asymmetry factors. Recursively reconstructing children from parents \add{follows the structure of an inverse lifting step}, so the resulting tree \add{can be read as} a \textit{wavelet decomposition} of the scene, not a level-of-detail index. We adopt the lifting scheme deliberately. When $\boldsymbol{\alpha}=\mathbf{1}$, the construction is the orthogonal Haar transform; $\boldsymbol{\alpha}\neq\mathbf{1}$ generalizes it to a biorthogonal transform, mirroring how JPEG2000~\cite{taubman2002jpeg} adopts biorthogonal CDF~9/7 filters over Haar to trade exact energy conservation for expressivity (detailed in \cref{sec:method:design_space}).

A child is its parent plus a detail, so per split we store only the detail $\boldsymbol{\psi}$ needed to reconstruct the children. 
Each split stores $D+5$ attribute parameters instead of $2D$ (\ie, two independent $D$-dimensional children). Also, because children are generated from their parent, they inherit its spatial support, and the tree topology encodes spatial locality by construction. This coherence is what makes the detail coefficients sparse (\cref{fig:energy_concentration}), and what lets viewpoint-aware and region-of-interest streaming reduce to a subtree traversal with no auxiliary spatial index.

Progressive training builds an adaptive refinement tree across increasing image resolutions. Finer levels can adjust the geometry represented by a coarse splat through its descendants, while the previously trained coarse level remains available for delivery. On the client, a tree cut defines the active splats, which are reconstructed by accumulating refinements and passed to the standard 3DGS rasterizer. This representation supports updates at individual splits as well as complete quality levels.

Our contributions are as follows:
\begin{itemize}
    \item \textit{A 3DGS representation for progressive delivery and scalable rendering.} We recast scalable 3DGS as a signal-decomposition problem and endow the unstructured Gaussian set with a critically-sampled and coarse-to-fine wavelet basis. This proposes a different paradigm from prior scalable 3DGS who stacks independent layers or merges redundant spatial pyramids.

    \item \textit{A lifting-scheme construction with biorthogonal refinement.} We generate finer splats from coarser ones by a learned lifting step whose per-attribute asymmetry factor generalizes the orthogonal Haar transform to a biorthogonal one, giving sparse detail coefficients that still capture non-symmetric detail.

    \item \textit{State-of-the-art scalable performance.} Extensive experiments show that {\name} cuts splat redundancy from over 65\% to under 25\%, reduces transmission payload and rendering GPU footprint by up to 2.4$\times$ and 5.5$\times$ over the strongest baselines, achieves smooth and monotonic quality improvements in progressive-delivery simulations, and compresses  8.7$\times$ relative to standard coding at only 0.10~dB loss. 
\end{itemize}

The remainder of this paper is organized as follows. \cref{sec:related} reviews the background and related work on 3DGS, multiscale 3DGS representations, and 3DGS compression techniques. \cref{sec:methodology} details the construction of our {\name}. \cref{sec:experiments} presents the experimental evaluations, including comparative results and a detailed analysis of inter-layer correlation. Finally, \cref{sec:conclusion} concludes the paper with future directions for improvement.

%% file: sec/2-related_work.tex
\section{Background and Related Work} \label{sec:related}

\subsection{3D Gaussian Splatting} \label{subsec:3dgs}

3DGS achieves real-time, photorealistic radiance-field rendering by representing a 3D scene as splats.
Each splat is defined as $G(\mathbf{x}) = \exp(-\tfrac{1}{2}\mathbf{x}^T \mathbf{\Sigma}^{-1} \mathbf{x})$ and characterized by a position $\mathbf{x}$, a covariance matrix $\mathbf{\Sigma}$ decomposed into a scaling matrix $\mathbf{S}$ and a rotation matrix $\mathbf{R}$, opacity $\sigma$, and Spherical Harmonic (SH) coefficients. 
At render time
, splats are projected into 2D camera coordinates and rasterized in tiles with $\alpha$-blending.
Initialized from a sparse Structure-from-Motion point cloud, the splats are then refined via gradient descent with adaptive densification.



\subsection{Multiscale 3DGS} \label{subsec:lod_3dgs}

Organizing 3DGS across scales and resolutions has been pursued from two perspectives with different goals: accelerating \textit{rendering}, and enabling \textit{streaming}.

\textbf{Hierarchies for rendering.}
One line partitions the scene spatially and selects a level per viewpoint at render time~\cite{kerbl2024hierarchical,ren2024octree,lu2024scaffold}. Hierarchical 3DGS~\cite{kerbl2024hierarchical} recursively merges neighboring splats into parent nodes, forming a tree whose cut is chosen per view to render large scenes in real time. Scaffold-GS~\cite{lu2024scaffold} anchors splats to voxels with MLP-predicted attributes, and Octree-GS~\cite{ren2024octree} organizes anchors into octree levels. In ~\cite{milef2025learning}, they learn a continuous level of detail that reorders splats by importance so that rendering a prefix yields a valid image. FLoD~\cite{seo2024flod} adds flexible discrete levels, and Mip-Splatting~\cite{yu2024mip} filters splats to render alias-free across scales.
These methods are effective for their goal, but the goal is render-time selection of a model that already resides on the client. The full hierarchy must be present to choose a cut. A coarser level is not a transmittable prefix of a finer one, and levels are stored in full. They are rendering accelerators, not progressive thus streamable representations.

\textbf{Quality layers for streaming.}
A second line targets progressive transmission: a base layer is sent first, and enhancement layers are added, so received prefix is renderable~\cite{shi2025lapis,tsai2025l3gs,sun2025lts,milef2025learning}. 
LapisGS~\cite{shi2025lapis} established this by training a base layer and enhancement layers at increasing resolutions, each a separate splat set. LTS~\cite{sun2025lts} builds a DASH-based streaming system on top of this representation, while L3GS~\cite{tsai2025l3gs} applies a similar paradigm at object-segmented granularity.
However, each layer is a global and independent set with no parameter-level relationship to coarser splats. Our experiments examine the resulting accumulation of transparent splats and uneven quality improvements during partial reception (\cref{sec:experiment:layer_analysis}). {\name} instead encodes a local parameter-level refinement relationship and refines a coarse splat, replacing it by its two children, when the detail coefficients have been received by the client.

\subsection{3DGS Compression} \label{subsec:compression_3dgs}

A parallel line reduces the size of 3DGS model. Pruning and vector quantization~\cite{fan2023lightgaussian}, and anchor-based entropy models such as HAC~\cite{chen2024hac,chen2025hacplus} and PCGS~\cite{chen2026pcgs}, achieve large size reductions, and a compressed full model can be smaller than a coarse streaming level.
Compression and multiresolution scalability are, however, orthogonal axes. 
These methods encode one quality level of one model into a latent form that must be decoded before rendering; they do not by themselves provide a scalable representation from which a client can request and render any quality tier. At render time, anchor-based codecs run an MLP per anchor for every quality update and materialize anchors into explicit splats, so the rendering footprint and per-frame cost remain high (HAC reports decoding time from 2s to 30s). \add{PCGS~\cite{chen2026pcgs} makes this pipeline progressive. Each further level refines the already decoded anchors by progressive quantization and adds new anchors, reusing the earlier bitstream. Every level, however, must still be entropy-decoded and converted into splats by the anchors' neural networks before it can be rendered. PCGS reports 4.9--20.5\,s to decode the first level and a further 0.8--3.4\,s per additional level. A {\name} client instead reconstructs a level with one multiply-add per attribute, in milliseconds (\cref{sec:experiment:client}), and passes the result to the standard rasterizer. Therefore, it is not trivial to apply compressed 3DGS for adaptive streaming.}
Nevertheless, compression methods and learned entropy models could complement this representation. Our quantization experiment examines the compressibility of its detail coefficients.

\subsection{Multiresolution and Wavelet Representations} \label{subsec:wavelets}

Our refinement draws on the classical theory of multiresolution analysis and wavelet lifting scheme~\cite{sweldens1996lifting}. In image and video coding, the wavelet transform is the standard route to scalable representations. JPEG2000~\cite{taubman2002jpeg} and embedded coders such as EZW~\cite{shapiro1993embedded} and SPIHT~\cite{said1996new} deliver resolution and quality scalability from a single embedded bitstream because a wavelet decomposition is multiresolution, critically sampled, and produces sparse detail coefficients. 
Wavelet ideas have also entered 3D representations: Masked-wavelet NeRF~\cite{rho2023masked} transforms a NeRF feature grid for compactness, and RAHT~\cite{de2016compression}, a Haar-like wavelet on octree-organized attributes, underlies point-cloud coding in the MPEG standard. These transform a regular domain (a grid or octree). 
\add{In 3DGS, Wavelet-GS~\cite{zhao2025wavelet} applies a 3D wavelet decomposition to the point cloud and a 2D wavelet decomposition to the training images, so that low- and high-frequency components are optimized separately. However, the decomposition aims at guiding training toward higher reconstruction quality, and the trained model is a single splat set with no multiresolution structure at delivery time. Our scope is different. We define a parent--child lifting step on the splats themselves, so the refinement tree is the delivered representation.}

%% file: sec/3-methodology.tex
\begin{figure}[t]
    \centering
    \includegraphics[width=0.95\columnwidth]{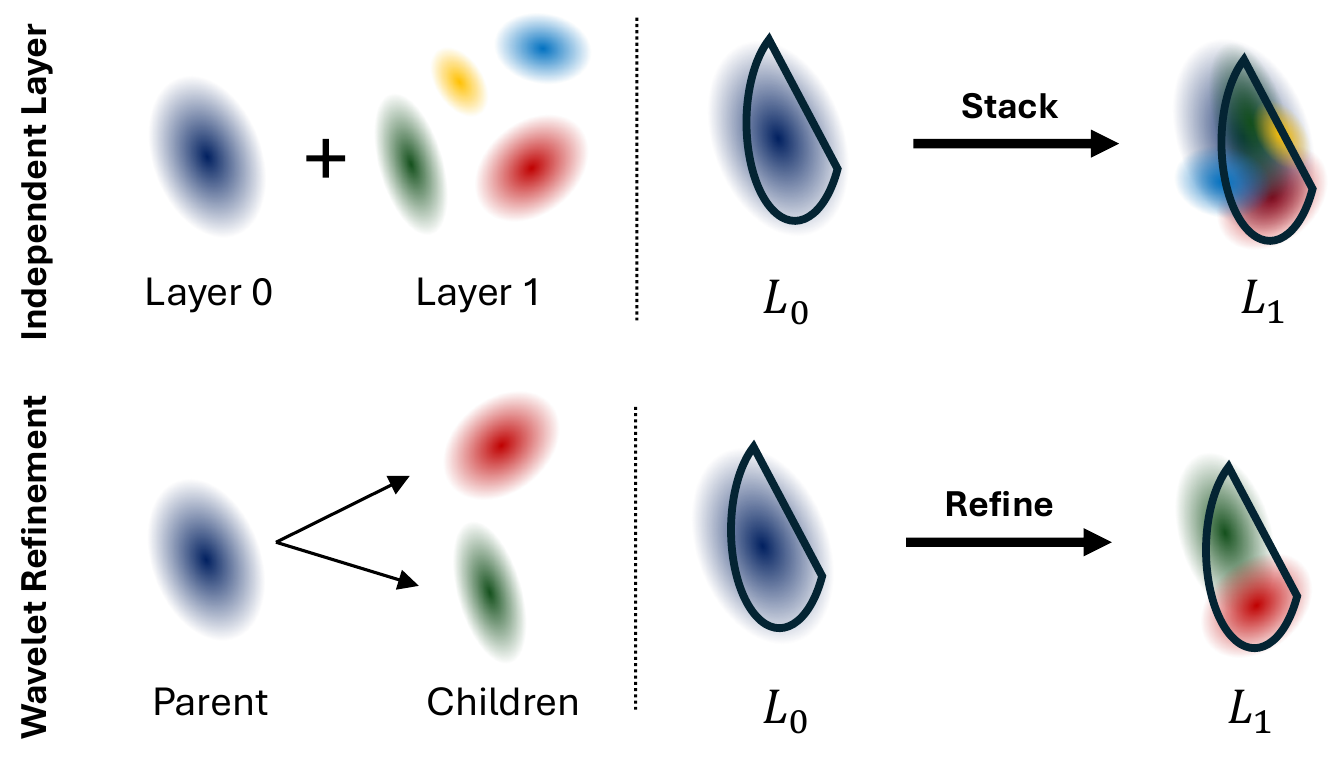}
    \caption{\textbf{Top (independent layers):} To fit the target geometry (black outline) at $L_1$, the splat from $L_0$ remains frozen and is always rendered at $L_1$; the quality transition is achieved by stacking a new splat on top.
    \textbf{Bottom (wavelet refinement):} The parent splat at $L_0$ refines into child splats that replace it at $L_1$. The children structurally adapt to fit finer geometry. 
    }
    \label{fig:discrete_vs_continuous}
\end{figure}

\section{Methodology} \label{sec:methodology}

The limitation of independent-layer streaming stems from a missing structural relationship: a splat at one scale has no parameter-level connection to its counterparts at other scales.
We resolve it by organizing the splats as a \textit{wavelet decomposition}, a tree in which each finer splat is generated from a coarser one through a learned detail coefficient.
We explain the parent-child parametrization, the tree structure, and the progressive pipeline below.

\subsection{Design Space: Parent-Child Refinement as Wavelet Lifting}
\label{sec:method:design_space}

Independent-layer redundancy stems from the lack of parameter-level connections across scales. Unlike spatial hierarchies (\eg, octrees) that partition space, our tree records how each fine splat is reconstructed from a coarse one: every leaf traces a unique path to a root, its parameters the accumulated detail along that path (\cref{fig:discrete_vs_continuous}). This gives the three streaming properties by design: (i) scales are related by an explicit, traversable transform; (ii) truncating the tree yields a valid scene with fewer redundant splats; and (iii) rendering cost and GPU memory scale with received data.
The central design question is how a child relates to its parent, \ie, the choice of transform.

\begin{figure}[t]
    \centering
    \begin{subfigure}[b]{0.45\columnwidth}
        \centering
        \includegraphics[width=\textwidth]{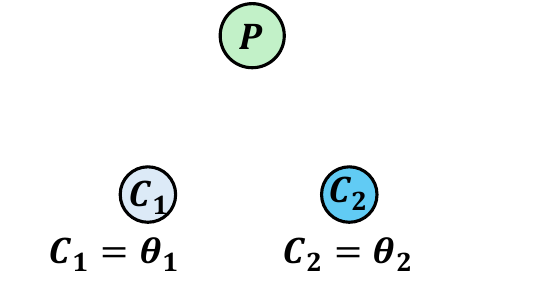}
        \caption{Option A ($2D$ param.).}
        \label{subfig:optionA}
    \end{subfigure}
    \begin{subfigure}[b]{0.45\columnwidth}
        \centering
        \includegraphics[width=\textwidth]{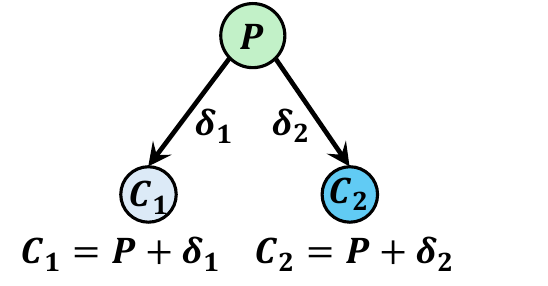}
        \caption{Option B ($2D$ param.).}
        \label{subfig:optionB}
    \end{subfigure}
    \\
    \begin{subfigure}[b]{0.45\columnwidth}
        \centering
        \includegraphics[width=\textwidth]{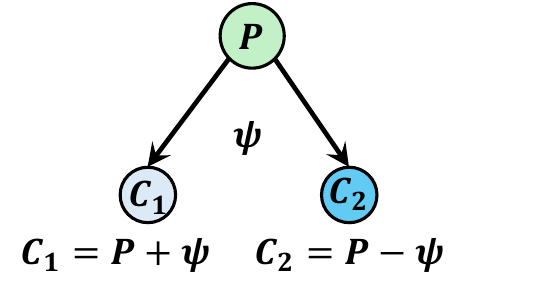}
        \caption{Option C ($D$ param.).}
        \label{subfig:optionC}
    \end{subfigure}
    \begin{subfigure}[b]{0.45\columnwidth}
        \centering
        \includegraphics[width=\textwidth]{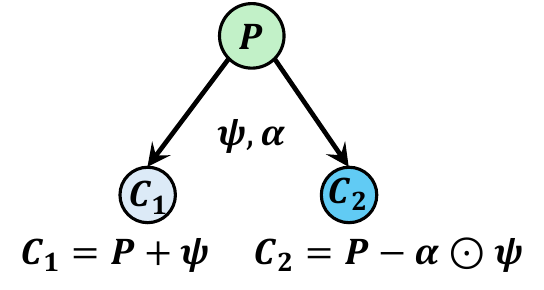}
        \caption{Option D ($D+5$ param.).}
        \label{subfig:optionD}
    \end{subfigure}
    \caption{\textbf{Design space for parent-child refinement.} 
    \textbf{(a, b)} Unconstrained constructions. 
    \textbf{(c)} A symmetric collinear residual (Haar wavelet) 
    is rigid for real-world details. \textbf{(d)} An asymmetric formulation relaxes the symmetry with a learned factor $\boldsymbol{\alpha}$, achieving high representational expressivity.}
    \label{fig:options}
\end{figure}

Let $\boldsymbol{P} \in \mathbb{R}^D$ denote the parameters of a parent splat, where $D$ encompasses position, rotation, scale, opacity, and SH coefficients. A binary split produces children $\boldsymbol{C_1}$ and $\boldsymbol{C_2}$. We evaluate four possible constructions (\cref{fig:options}).

\textbf{Option A: Independent children.} 
Each child is parameterized independently:
\begin{equation}
    \boldsymbol{C_1} = \boldsymbol{\theta_1}, \qquad \boldsymbol{C_2} = \boldsymbol{\theta_2},
\end{equation}
where $\boldsymbol{\theta_1}, \boldsymbol{\theta_2} \in \mathbb{R}^D$ are freely optimized. This is equivalent to the densification strategy of standard 3DGS~\cite{kerbl20233d}: a splat splits into two that are then optimized without constraint. 
The resulting tree is, in effect, an independent-layer representation that merely records the splitting history, inheriting all its limitations.
The parent $\boldsymbol{P}$ bears no guaranteed relationship to $\boldsymbol{\theta_1}$ or $\boldsymbol{\theta_2}$, so coarse rendering is unreliable, violating \textit{rendering scalability}. With $2D$ uncorrelated parameters per split, the representation also offers nothing for \textit{incremental adaptation} to exploit.

\textbf{Option B: Independent residuals.}
Each child is defined as an independent offset from the parent:
\begin{equation}
    \boldsymbol{C_1} = \boldsymbol{P} + \boldsymbol{\delta_1}, \qquad \boldsymbol{C_2} = \boldsymbol{P} + \boldsymbol{\delta_2},
\end{equation}
where $\boldsymbol{\delta_1}, \boldsymbol{\delta_2} \in \mathbb{R}^D$ are freely optimized. Mathematically, this is Option A with a different initialization that biases children toward the parent at the early training, 
but the offsets can grow in arbitrary directions as training progresses. 
The two offsets remain uncorrelated, the parameter cost remains $2D$ per split, and the limitations for \textit{rendering scalability} and \textit{incremental adaptation} persist. 

Options A and B reveal a clear pattern: without structural enforcement of parent-child proximity, coarse-depth rendering quality is unpredictable and the refinement data is unstructured. 
Also, both approaches are highly inefficient for transmission, as they require $2D$ parameters per split. 
Meeting the streaming principles demands a stronger constraint on how children relate to the parent. 


\textbf{Option C: Symmetric collinear residual.}
A constraint requires children to deviate from parent along a shared direction equally:
\begin{equation}
    \boldsymbol{C_1} = \boldsymbol{P} + \boldsymbol{\psi}, \qquad \boldsymbol{C_2} = \boldsymbol{P} - \boldsymbol{\psi}, \qquad \frac{\boldsymbol{C_1} + \boldsymbol{C_2}}{2} = \boldsymbol{P},
\end{equation}
where $\boldsymbol{\psi} \in \mathbb{R}^D$ is a learned refinement vector. This halves the parameter cost to $D$ per split.
Rendering the parent in place of its two unsent children recovers their exact mean, \ie the optimal $L^2$ approximation at that scale, making coarse-depth rendering provably optimal under the mean-square criterion, directly satisfying \textit{rendering scalability}.

\begin{figure}[t]
    \centering
    \begin{subfigure}[b]{0.49\columnwidth}
        \centering
        \includegraphics[width=\textwidth]{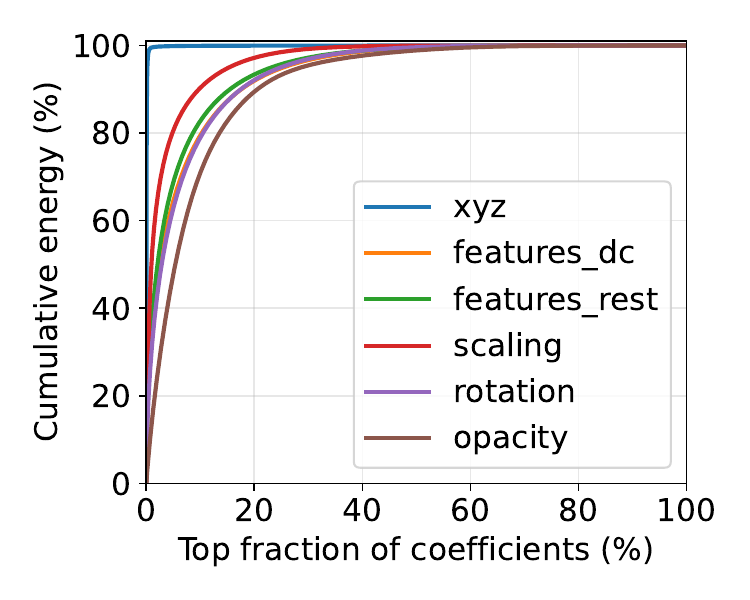}
        \caption{Symmetric (Option C).}
        \label{subfig:energy_concentration_sym}
    \end{subfigure}
    \hfill 
    \begin{subfigure}[b]{0.49\columnwidth}
        \centering
        \includegraphics[width=\textwidth]{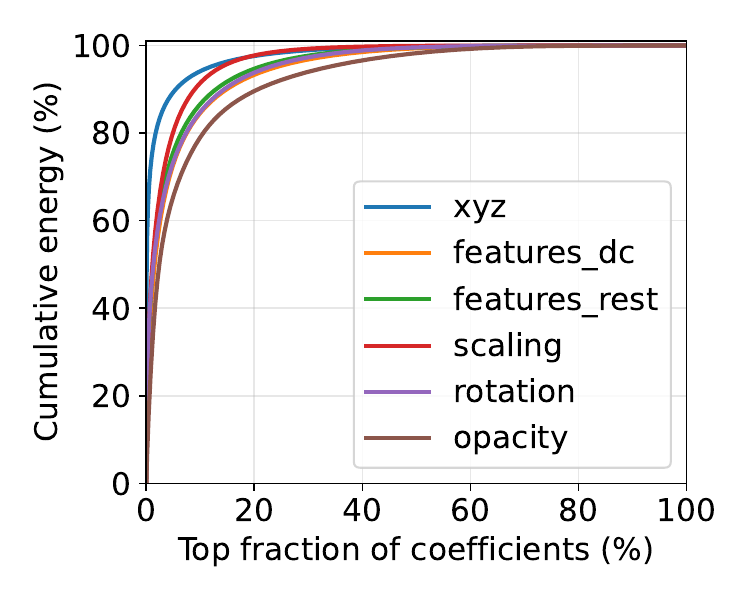}
        \caption{Asymmetric (Option D).}
        \label{subfig:energy_concentration_asym}
    \end{subfigure}
    \caption{\textbf{Cumulative energy concentration of $\boldsymbol{\psi}$.} Both constructions retain wavelet sparsity: fewer than 20\% of coefficients carry over 90\% of the energy.}
    \label{fig:energy_concentration}
\end{figure}

This is the Haar wavelet lifting scheme~\cite{sweldens1996lifting}: $\boldsymbol{P}$ is the approximation coefficient and $\boldsymbol{\psi}$ the detail coefficient, and recursively reconstructing children from parents is an inverse discrete wavelet transform. Thus $\boldsymbol{\psi}$ is sparse and near-zero for locally smooth signals (\cref{subfig:energy_concentration_sym}), so the symmetric construction satisfies \textit{rendering scalability}, yields compact signals favorable for \textit{incremental adaptation}, and inherits wavelet sparsity guarantees.
However, this rigid symmetry wastes capacity on real scenes. For instance, at a foreground-background boundary, one child must move substantially to capture foreground texture while the other should remain static for the smooth background. Forced symmetry over-refines the background or under-refines the foreground, penalizing \textit{quality scalability} (validated in \cref{sec:experiment:eva,sec:experiment:layer_analysis}).

\textbf{Option D: Asymmetric collinear residual.}
To recover expressivity, we relax the symmetric constraint with a learned per-attribute asymmetry vector $\boldsymbol{\alpha} \in \mathbb{R}^A$ ($A=5$ for position, rotation, scale, opacity, and SH):
\begin{equation}
    \boldsymbol{C_1} = \boldsymbol{P} + \boldsymbol{\psi},  \boldsymbol{C_2} = \boldsymbol{P} - \boldsymbol{\alpha} \odot \boldsymbol{\psi},  \frac{\boldsymbol{C_1} + \boldsymbol{C_2}}{2} = \boldsymbol{P}  + \frac{(\mathbf{1}-\boldsymbol{\alpha}) \odot \boldsymbol{\psi}}{2},
    \label{eq:asymmetric}
\end{equation}
requiring $D + 5$ parameters per split. The collinearity constraint is preserved, as the children still lie along a single axis through $\boldsymbol{P}$.

Like JPEG2000's~\cite{taubman2002jpeg} move from orthogonal Haar to biorthogonal filters, $\boldsymbol{\alpha}$ trades exact mean preservation for expressivity while retaining the wavelet framework's essentials. To verify this empirically, we construct trees for Options C and D on synthetic Blender~\cite{mildenhall2021nerf}, and measure the cumulative energy concentration of $\boldsymbol{\psi}$ (\cref{fig:energy_concentration}). As observed, both Option~C and D keep $\boldsymbol{\psi}$ tightly clustered near zero, with fewer than 20\% of coefficients carrying over \SI{90}{\percent} of the energy (\cref{fig:energy_concentration}); the asymmetric form thus retains the sparsity that makes wavelet coding effective (exploited in \cref{sec:experiment:layer_analysis}). We therefore adopt Option~D.

\textbf{Tree structure.}
The wavelet decomposition tree consists of \textit{root nodes} (baseline approximation $\boldsymbol{P_{\text{root}}} \in \mathbb{R}^D$), \textit{internal nodes} (storing learned detail coefficients $\boldsymbol{\psi} \in \mathbb{R}^D$ and asymmetry $\boldsymbol{\alpha} \in \mathbb{R}^A$), and \textit{leaf nodes} (renderable splats).
A leaf's final parameters $\boldsymbol{S}$ are computed by accumulating refinements from root to leaf: 
\begin{equation}
    \boldsymbol{S} = \boldsymbol{P_{\text{root}}} + \sum_{k=1}^{\ell} \boldsymbol{s_k} \odot \boldsymbol{\psi_k},
    \label{eq:leaf_reconstruction}
\end{equation}
where $\ell$ is leaf depth and $\boldsymbol{s_k} \in \{+\mathbf{1},\; -\boldsymbol{\alpha}_k\}$ indicates the branch taken.
A node's role is dynamic: at a coarse level it is a renderable leaf, at a finer level an internal node passing parameters to its children, so a single stored tree serves all quality tiers.

\subsection{Progressive Training}
\label{sec:method:training}

\textbf{Multi-level optimization.}
We use $N{+}1$ quality levels $L_0,\ldots,L_N$, one per training stage. From a full-resolution multi-view set $\mathbf{D}_N$ we build an image pyramid $\{\mathbf{D}_i\}_{i=0}^{N}$:
    $\mathbf{D}_i = \left\{\left(\mathbf{V_m^i}, \mathbf{X_m^i}\right)\right\}_{m=1}^M,$
where $\mathbf{V_m^i}$ is the camera matrix, $\mathbf{X_m^i}$ is the corresponding image at the $i$-th resolution, and $M$ is the number of views.

\textit{Root initialization and $L_0$ refinement.}
We first train a standard 3DGS base on $\mathbf{D}_0$ to initialize the root set $\mathcal{R}$. We then apply lifting-based refinement on the same image set: selected splats (\add{whose positional gradients exceed a threshold}) are split into children through the learned $\boldsymbol{\psi}$ and $\boldsymbol{\alpha}$ parameterization. The active leaves after this refinement phase form $\mathcal{F}_0$, the representation rendered at quality level $L_0$. Thus, $\mathcal{R}$ is the initialization of the refinement tree, while $\mathcal{F}_0$ already incorporates learned refinements.

\textit{Higher quality levels.}
The leaf set $\mathcal{F}_i$ is all leaves after level-$L_i$ training; rendering at $L_i$ rasterizes exactly $\mathcal{F}_i$. At each stage, a subset of leaves is replaced by finer descendants to produce
$
    \mathcal{F}_{i+1} = 
    \bigl(\mathcal{F}_i \setminus \mathcal{S}_i\bigr) 
    \;\cup\; \mathcal{C}(\mathcal{S}_i),
    \label{eq:leaf_set_transition}
$
where $\mathcal{S}_i \subseteq \mathcal{F}_i$ is the set of leaves that split during level-$L_{i+1}$ training, and $\mathcal{C}(\mathcal{S}_i)$ denotes the resulting terminal descendants, obtained through one or more splits via \cref{eq:asymmetric}. 
Leaves not selected for splitting (because their local geometry is already well-represented) persist unchanged into $\mathcal{F}_{i+1}$. 

At each subsequent level $L_i$ ($i\geq1$), the leaf set $\mathcal{F}_{i-1}$ serves as the trainable frontier. For each node $n \in \mathcal{F}_{i-1}$, its refinement direction $\boldsymbol{\psi}_n$ and asymmetry vector $\boldsymbol{\alpha}_n$ are unfrozen, while ancestor parameters remain fixed. 
Because children are defined relative to their parent via \cref{eq:asymmetric}, gradient updates to $\boldsymbol{\psi}$ and $\boldsymbol{\alpha}$ refine the scene by correcting the parent representation, rather than masking its deficiencies with independent splats.

The optimization minimizes a rendering loss $\mathcal{L}_i$ consisting of an $L_1$-norm term and a D-SSIM term:
$
    \mathcal{L}_i = (1-\lambda) \sum_{m=1}^{M} 
    \mathcal{L}_1(\mathbf{\hat{X}_m^i}, \mathbf{X_m^i}) + \lambda 
    \sum_{m=1}^{M} \mathcal{L}_{\text{D-SSIM}}(\mathbf{\hat{X}_m^i}, 
    \mathbf{X_m^i}),
    \label{eq:loss}
$
where $X_m^i$ is the ground truth for view $m$ at the $i$-th quality level, $\hat{X}_m^i$ is the image rendered from the current leaf set, and $\lambda = 0.2$ following the default setting~\cite{kerbl20233d}.

\textbf{Adaptive densification.}
Following standard 3DGS~\cite{kerbl20233d}, leaves whose accumulated 2D positional gradients exceed a threshold \add{($2\times10^{-4}$)} split into two children. Because splitting is gradient-driven, the tree is unbalanced and spatially adaptive, and a leaf may split recursively within one stage.
Once level $L_i$ converges, its frontier $\boldsymbol{\psi}$ and $\boldsymbol{\alpha}$ are frozen as $\mathcal{F}_i$ and training advances to $L_{i+1}$, repeating to $L_N$; the leaf sets $\mathcal{F}_0, \ldots, \mathcal{F}_N$ form the $N{+}1$ quality tiers. 
For streaming, the server first sends the root set $\mathcal{R}$, followed by the refinement parameters $\boldsymbol{\psi}$ and $\boldsymbol{\alpha}$ learned within $L_0$. Together, these reconstruct the active leaf set $\mathcal{F}_0$ for rendering at $L_0$. Subsequent payloads contain the additional refinement parameters needed to reconstruct $\mathcal{F}_1, \ldots, \mathcal{F}_N$, including all recursive splits within each stage.
Because the tree grows adaptively, $\mathcal{F}_i$ holds more splats in complex regions and fewer in smooth ones, so rendering cost tracks the scene's detail distribution rather than a uniform resolution.

\begin{table*}[t!]
    \caption{
        Quantitative comparison on four datasets (Blender~\cite{mildenhall2021nerf}, Mip-NeRF360~\cite{barron2022mip}, Tank\&Temples~\cite{knapitsch2017tanks}, Deep Blending~\cite{hedman2018deep}) at four quality levels. Stor.: cumulative transmitted data (MB). Mem.: GPU memory the rasterizer needs, i.e., the materialized leaf set (MB). $\Delta$: difference between the best and the second-best methods. In each cell the best result is shown in bold. \add{Each level is evaluated at its own resolution. Blender scores are higher because its scenes are synthetic objects on clean backgrounds, consistent with prior work~\cite{kerbl20233d}.} SSIM and LPIPS follow the same trends \add{and can be explored per scene on the project website}.
    }
    \label{tab:quantitative_results_all}
    \centering
    \small
    \setlength{\tabcolsep}{5pt}
        \begin{tabular}{cc|ccc|ccc|ccc|ccc}
            \hline
            \multirow{2}{*}{} & \multirow{2}{*}{Method} & \multicolumn{3}{c|}{$L_0$} & \multicolumn{3}{c|}{$L_1$} & \multicolumn{3}{c|}{$L_2$} & \multicolumn{3}{c}{$L_3$} \\
            & & PSNR$\uparrow$ & Stor.$\downarrow$ & Mem.$\downarrow$ & PSNR$\uparrow$ & Stor.$\downarrow$ & Mem.$\downarrow$ & PSNR$\uparrow$ & Stor.$\downarrow$ & Mem.$\downarrow$ & PSNR$\uparrow$ & Stor.$\downarrow$ & Mem.$\downarrow$ \\
            \hline
            \multirow{5}{*}{\rotatebox{90}{Blender}}
            & Mono. & 36.19 & 18.96 & 18.96 & 35.60 & 39.09 & 39.09 & 34.98 & 75.63 & 75.63 & 33.72 & 147.13 & 147.13 \\
            & Single & 19.32 & 147.13 & 147.13 & 22.81 & 147.13 & 147.13 & 27.93 & 147.13 & 147.13 & 33.72 & 147.13 & 147.13 \\
            \cline{2-14}
            & L3GS  & 36.23 & 36.78 & 36.78 & 35.30 & 73.57 & 73.57 & 33.76 & 110.35 & 110.35 & 32.31 & 147.13 & 147.13 \\
            & LapisGS  & 36.18 & \bfseries18.98 & 18.98 & 34.95 & 38.86 & 38.86 & 33.36 & 73.78 & 73.78 & 32.52 & 133.90 & 133.90 \\
            & {\name}  & \bfseries36.29 & 21.25 & \bfseries15.14 & \bfseries35.35 & \bfseries35.04 & \bfseries16.37 & \bfseries34.10 & \bfseries49.01 & \bfseries19.53 & \bfseries32.71 & \bfseries62.59 & \bfseries23.97 \\
            & $\Delta$  & \better{+0.06} & +10.7\% & \better{-20.2\%} & \better{+0.05} & \better{-9.8\%} & \better{-57.9\%} & \better{+0.34} & \better{-33.6\%} & \better{-73.5\%} & \better{+0.19} & \better{-53.3\%} & \better{-82.1\%} \\
            \hline
            \hline
            \multirow{5}{*}{\rotatebox{90}{Mip-360}}
            & Mono. & 28.73 & 239.47 & 239.47 & 28.42 & 405.28 & 405.28 & 27.98 & 577.87 & 577.87 & 27.39 & 679.13 & 679.13 \\
            & Single & 19.86 & 679.13 & 679.13 & 22.69 & 679.13 & 679.13 & 26.16 & 679.13 & 679.13 & 27.39 & 679.13 & 679.13 \\
            \cline{2-14}
            & L3GS  & 28.72 & 169.78 & 169.78 & 28.20 & 339.57 & 339.57 & 27.31 & 509.35 & 509.35 & 26.15 & 679.13 & 679.13 \\
            & LapisGS  & 28.73 & 240.52 & 240.52 & 28.23 & 561.04 & 561.04 & 27.43 & 964.08 & 964.08 & \bfseries26.68 & 1382.05 & 1382.05 \\
            & {\name}  & \bfseries29.04 & \bfseries157.83 & \bfseries123.20 & \bfseries28.69 & \bfseries302.33 & \bfseries200.52 & \bfseries27.73 & \bfseries464.12 & \bfseries298.84 & 26.65 & \bfseries606.73 & \bfseries385.25 \\
            & $\Delta$  & \better{+0.31} & \better{-7.0\%} & \better{-27.4\%} & \better{+0.46} & \better{-11.0\%} & \better{-40.9\%} & \better{+0.30} & \better{-8.9\%} & \better{-41.3\%} & +0.03 & \better{-10.7\%} & \better{-43.3\%} \\
            \hline
            \hline
            \multirow{5}{*}{\rotatebox{90}{T\&T}}
            & Mono. & 24.45 & 87.57 & 87.57 & 24.55 & 149.97 & 149.97 & 24.31 & 256.02 & 256.02 & 23.71 & 434.18 & 434.18 \\
            & Single & 17.33 & 434.18 & 434.18 & 19.67 & 434.18 & 434.18 & 22.52 & 434.18 & 434.18 & 23.71 & 434.18 & 434.18 \\
            \cline{2-14}
            & L3GS  & 25.09 & 108.55 & 108.55 & 24.72 & 217.09 & 217.09 & 23.85 & 325.63 & 325.63 & 22.56 & 434.18 & 434.18 \\
            & LapisGS  & 24.62 & \bfseries86.85 & 86.85 & 24.58 & 208.42 & 208.42 & 24.13 & 394.19 & 394.19 & \bfseries23.42 & 691.44 & 691.44 \\
            & {\name}  & \bfseries25.57 & 88.45 & \bfseries77.89 & \bfseries25.41 & \bfseries186.82 & \bfseries94.26 & \bfseries24.67 & \bfseries270.18 & \bfseries119.52 & \bfseries23.42 & \bfseries354.84 & \bfseries149.70 \\
            & $\Delta$  & \better{+0.48} & +1.8\% & \better{-10.3\%} & \better{+0.69} & \better{-10.4\%} & \better{-54.8\%} & \better{+0.54} & \better{-17.0\%} & \better{-63.3\%} & {+0.00} & \better{-18.3\%} & \better{-65.5\%} \\
            \hline
            \hline
            \multirow{5}{*}{\rotatebox{90}{DB}}
            & Mono. & 27.01 & 261.79 & 261.79 & 28.54 & 419.81 & 419.81 & 29.43 & 564.46 & 564.46 & 29.56 & 665.59 & 665.59 \\
            & Single & 24.54 & 665.59 & 665.59 & 27.38 & 665.59 & 665.59 & 29.33 & 665.59 & 665.59 & 29.56 & 665.59 & 665.59 \\
            \cline{2-14}
            & L3GS  & 27.24 & 166.40 & 166.40 & 28.27 & 332.79 & 332.79 & 28.79 & 499.19 & 499.19 & 28.69 & 665.59 & 665.59 \\
            & LapisGS  & 27.02 & 262.79 & 262.79 & 28.29 & 493.64 & 493.64 & 29.00 & 744.68 & 744.68 & 28.97 & 999.82 & 999.82 \\
            & {\name}  & \bfseries27.78 & \bfseries133.10 & \bfseries101.35 & \bfseries28.81 & \bfseries230.72 & \bfseries139.19 & \bfseries29.30 & \bfseries323.18 & \bfseries179.74 & \bfseries29.10 & \bfseries405.30 & \bfseries215.43 \\
            & $\Delta$  & \better{+0.54} & \better{-20.0\%} & \better{-39.1\%} & \better{+0.52} & \better{-30.7\%} & \better{-58.2\%} & \better{+0.30} & \better{-35.3\%} & \better{-64.0\%} & \better{+0.13} & \better{-39.1\%} & \better{-67.6\%} \\
            \hline
        \end{tabular}
\end{table*}

\subsection{Rendering}
\label{sec:method:rendering}

At render time, the client requests a quality level $L_i$. The parameters of the active leaf set $\mathcal{F}_i$ are reconstructed via \cref{eq:leaf_reconstruction} and passed directly to the tile-based rasterizer. 

\textbf{Inherently smooth quality transitions.}
Independent-layer streaming ``pops'' when switching levels, which prior methods patch by interpolating adjacent splats at render time~\cite{shi2025lapis,kerbl2024hierarchical}. In {\name}, moving from $\mathcal{F}_i$ to $\mathcal{F}_{i+1}$ replaces parents with children offset by a sparse $\boldsymbol{\psi}$, so incremental updates are minute and localized: smoothness is a property of the representation, not a post-hoc trick (\cref{fig:smooth_transition,fig:sweep_psnr_lego_demo}). Detailed analysis and quantification is in \cref{sec:experiment:layer_analysis}.

\textbf{Per-region adaptive rendering.}
Because our model is traversed as a tree, spatial regions can be rendered at mixed resolutions simultaneously. For example, in bandwidth-constrained VR scenarios, foveal regions can be traversed to $L_N$ while peripheral regions stop at $L_j$ ($j < N$). The rasterizer simply renders the mixed leaf set 
$\mathcal{F}_N^{\text{foveal}} \cup \mathcal{F}_j^{\text{periph.}}$ in a single pass. This fine-grained spatial adaptation is structurally impossible in independent-layering representations.

%% file: sec/4-experiment.tex
\begin{figure*}[t!]%
    \centering
    \begin{subfigure}{0.245\linewidth}
        \includegraphics[width=\linewidth]{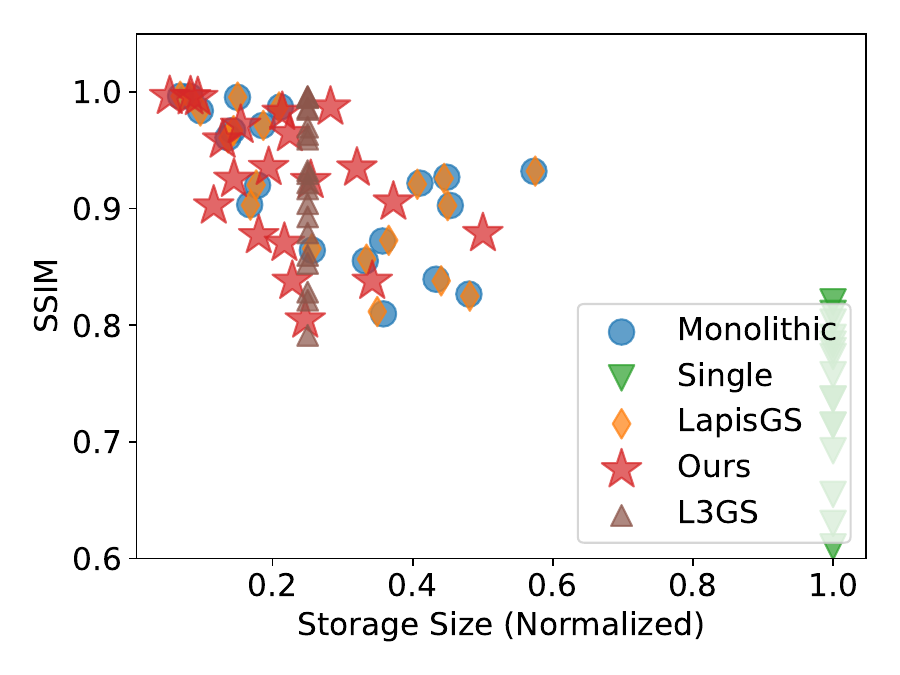}
        \caption{$L_0$.}
    \end{subfigure}
    \begin{subfigure}{0.245\linewidth}
        \includegraphics[width=\linewidth]{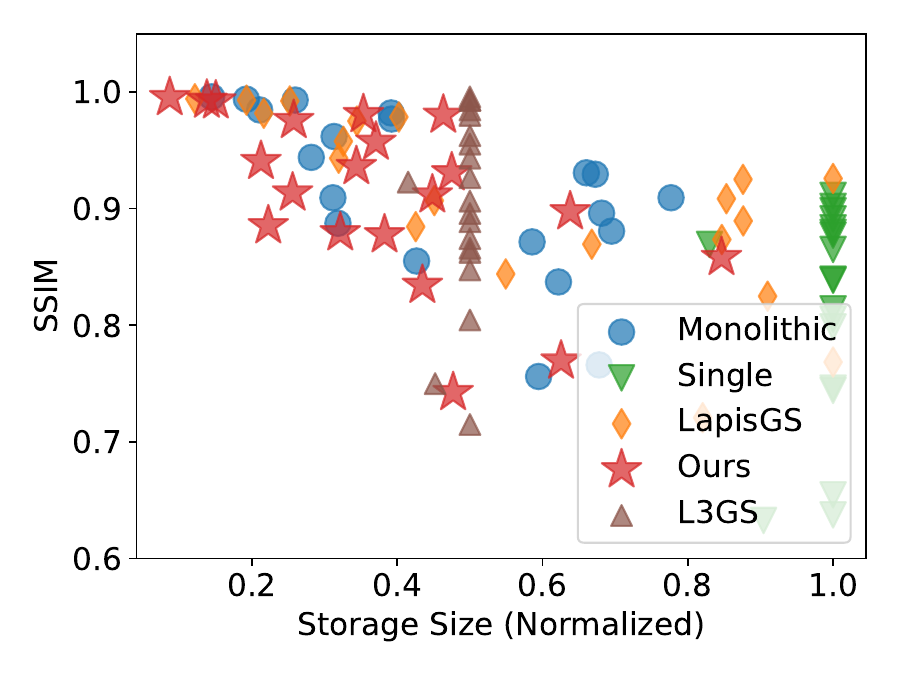}
        \caption{$L_1$.}
    \end{subfigure}
    \begin{subfigure}{0.245\linewidth}
        \includegraphics[width=\linewidth]{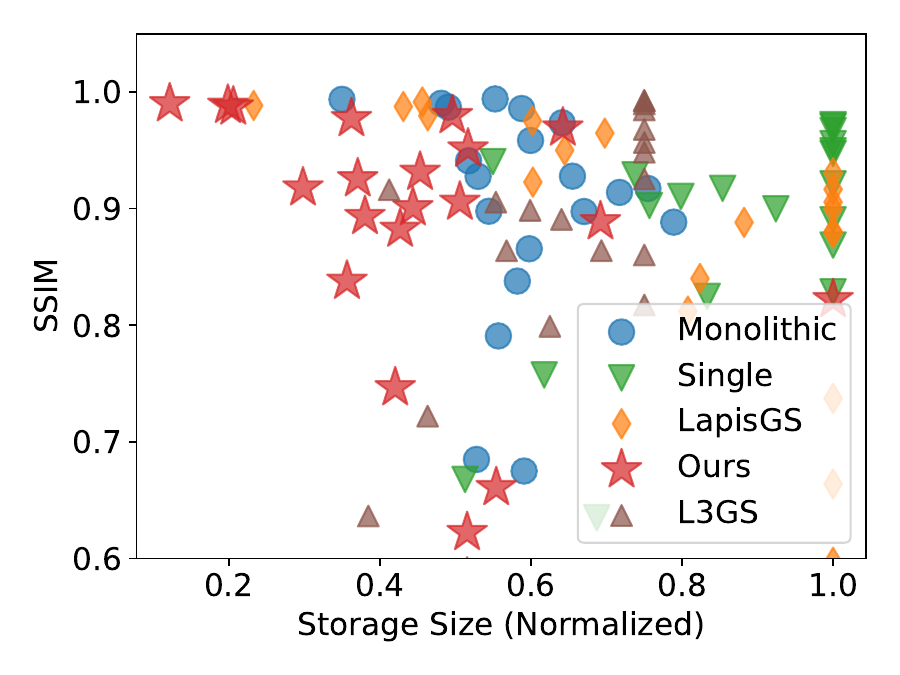}
        \caption{$L_2$.}
    \end{subfigure}
    \begin{subfigure}{0.245\linewidth}
        \includegraphics[width=\linewidth]{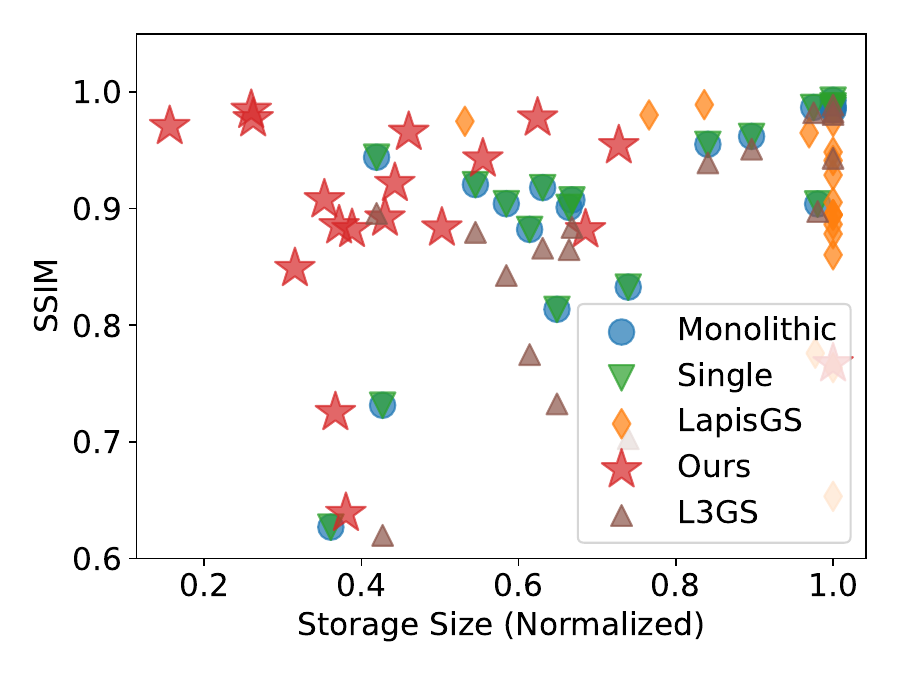}
        \caption{$L_3$.}
    \end{subfigure}
    \caption{\textbf{Quality-storage tradeoff for each scene at each quality level}. Each point represents the overall quality of a scene/object with the corresponding storage size at a given quality level.
    }%
    \label{fig:scatter_storage}
\end{figure*}

\begin{figure*}[t!]%
    \centering
    \begin{subfigure}{0.245\linewidth}
        \includegraphics[width=\linewidth]{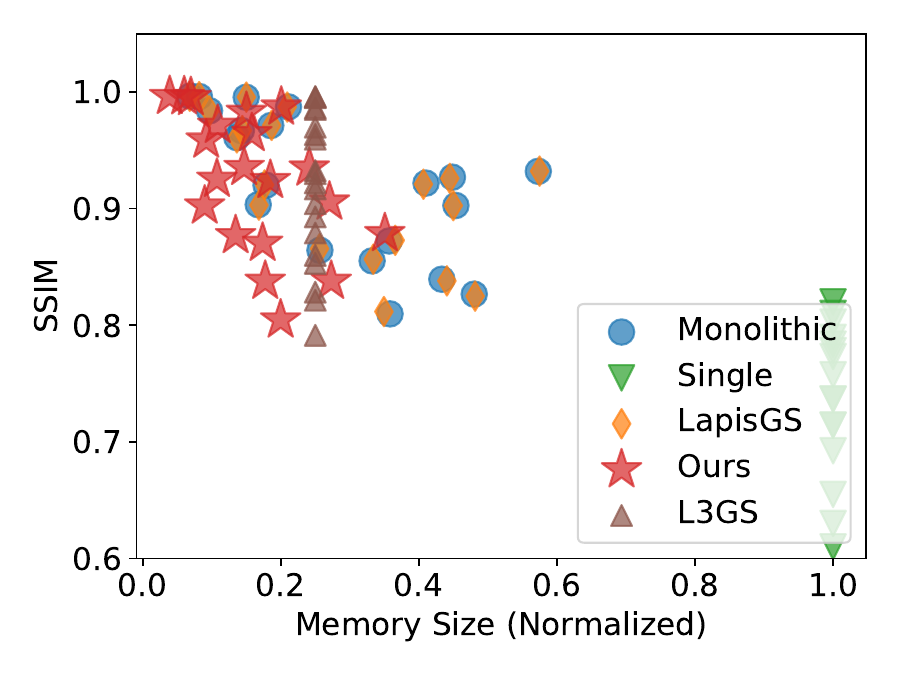}
        \caption{$L_0$.}
    \end{subfigure}
    \begin{subfigure}{0.245\linewidth}
        \includegraphics[width=\linewidth]{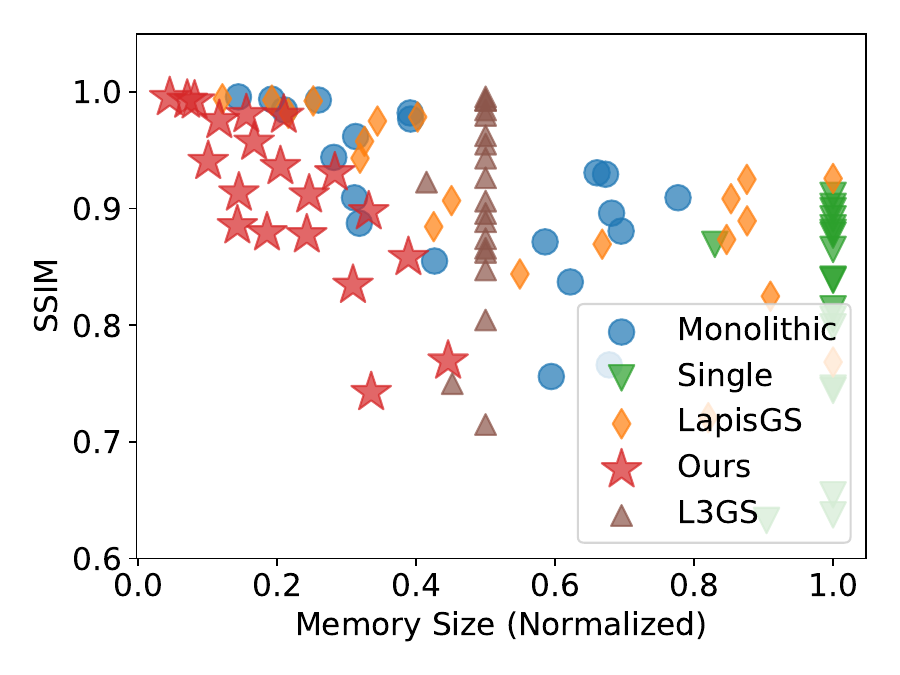}
        \caption{$L_1$.}
    \end{subfigure}
    \begin{subfigure}{0.245\linewidth}
        \includegraphics[width=\linewidth]{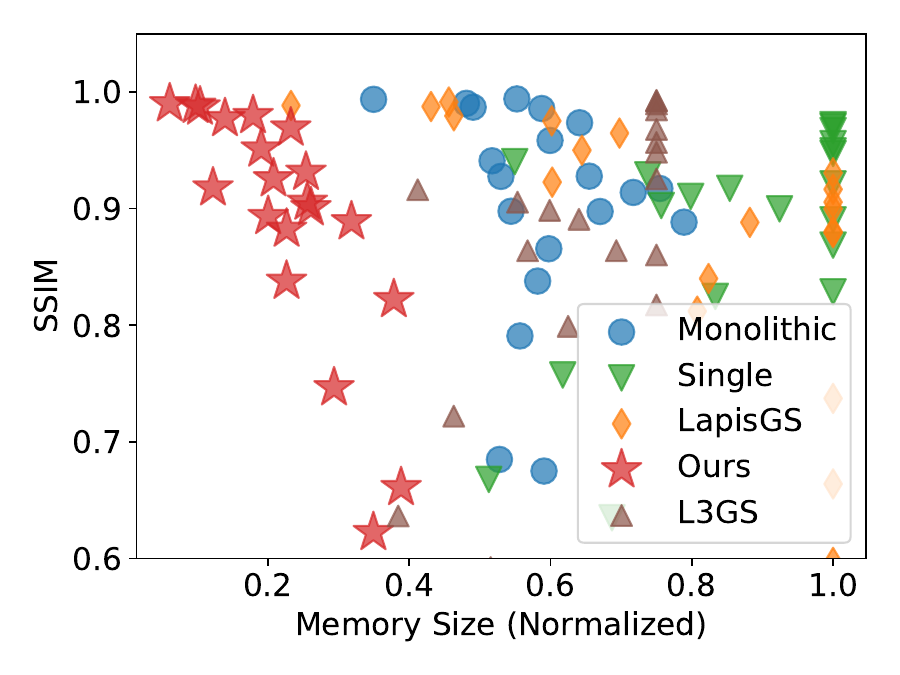}
        \caption{$L_2$.}
    \end{subfigure}
    \begin{subfigure}{0.245\linewidth}
        \includegraphics[width=\linewidth]{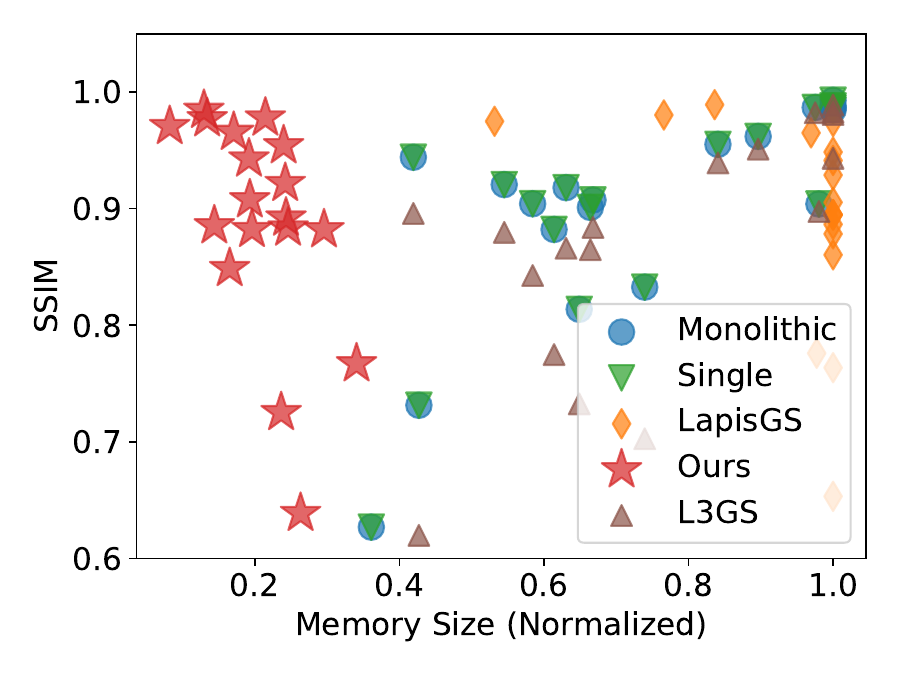}
        \caption{$L_3$.}
    \end{subfigure}
    \caption{\textbf{Quality-memory tradeoff for each scene at each quality level}. Each point represents the overall quality of a scene/object with the corresponding memory size at a given quality level.
    }%
    \label{fig:scatter_memory}
\end{figure*}

\section{Experiments} \label{sec:experiments}

We evaluate {\name} to validate the advantages of the wavelet decomposition. \cref{sec:experiment:setup} details the experimental setup. \cref{sec:experiment:eva} assesses the tradeoff among visual fidelity, transmission storage, and rendering footprint, and \cref{sec:experiment:layer_analysis} analyzes inter-level correlation, showing how the wavelet decomposition eliminates splat redundancy, smooths quality transitions, and yields compressible refinement signals. \add{\Cref{sec:experiment:client} measures the client-side overhead of applying refinements and rendering.}

\subsection{Experimental Setup} \label{sec:experiment:setup}

\textbf{Dataset.} 
We evaluate on all scenes from the standard 3DGS benchmark~\cite{kerbl20233d} across Synthetic Blender~\cite{mildenhall2021nerf}, Mip-NeRF360 (Mip-360)~\cite{barron2022mip}, Tanks\&Temples (T\&T)~\cite{knapitsch2017tanks}, and Deep Blending (DB)~\cite{hedman2018deep}. 
To construct the image pyramid $\{\mathbf{D}_0, \mathbf{D}_1, \mathbf{D}_2, \mathbf{D}_3\}$ required for progressive training, we downsample full-resolution images by factors of $8\times$, $4\times$, and $2\times$.

\textbf{Comparison methods.} 
We compare {\name} against:
    i) \textit{Monolithic.} A separate, independent 3DGS model trained from scratch \add{for 30k iterations} at each quality level. 
    This serves as the quality upper bound, which is the best possible rendering without streaming constraint.
    ii) \textit{Single.} One full-resolution 3DGS model trained on $\mathbf{D}_3$ and rendered at all resolution levels without per-level optimization.
    This represents the lower bound where no adaptation to different quality levels is performed. 
    iii) \textit{LapisGS~\cite{shi2025lapis}.} The state-of-the-art independent-layer method, training enhancement layers at increasing resolutions, each a separate splat set. 
    iv) \textit{L3GS~\cite{tsai2025l3gs}.} A recent independent-layer method defining layers by global significance score buckets. We adapt it to multi-resolution by training each layer on the corresponding pyramid resolution with dynamic opacity optimization, while preserving its pruning-based splat count control, GSG-sorted partitioning, and frozen ancestors.
    v) \textit{{\name} (Symmetric).} An ablation where the asymmetry factor is fixed to $\boldsymbol{\alpha} = \mathbf{1}$, reducing the tree to a symmetric Haar wavelet construction. 

\textbf{Metrics.}
We evaluate rendering quality using PSNR, SSIM, and LPIPS. \add{Unless noted otherwise, each level $L_i$ is rendered at the resolution of $\mathbf{D}_i$ and compared with the ground truth at that resolution, so scores are not comparable across levels.} We also report two size metrics that capture distinct aspects of the streaming and rendering pipeline: 
\textit{Storage size} (Stor.) measures the cumulative transmission payload required to reconstruct a given quality level. \textit{Rendering footprint} (Mem.) measures the GPU memory the rasterizer needs, i.e., the splats it reads to render a frame. For baselines, storage size always equals rendering footprint. For {\name}, the rendering footprint is smaller\add{, as the rasterizer reads only the materialized leaf set:} internal node parameters ($\boldsymbol{\psi}$, $\boldsymbol{\alpha}$) \add{are used to construct the leaves (\cref{sec:experiment:client}) but are not read during rendering}.

\textbf{Implementation details.}
Our implementation is built on the official codebases~\cite{kerbl20233d,shi2025lapis,tsai2025l3gs} and executed on an NVIDIA H100 GPU. 
We initialize the roots with 15k iterations of standard 3DGS on $\mathbf{D}_0$. We then train the lifting-based representation for 30k iterations at each quality level, including $L_0$, using the corresponding $\mathbf{D}_i$. Thus, obtaining $\mathcal{F}_0$ uses 45k iterations in total, while each subsequent level adds 30k iterations. Within each 30k-iteration refinement stage, densification runs every 3k iterations through iteration 15k.
All other hyperparameters match default 3DGS settings.

\subsection{Visual Quality v.s. Storage Size} \label{sec:experiment:eva}

\textbf{Quantitative results.} \Cref{tab:quantitative_results_all} reports the results on four datasets. \Cref{fig:scatter_storage,fig:scatter_memory} visualize the same comparison per scene: each point is one scene at one quality level, plotted as SSIM against normalized storage or memory, and points toward the upper left indicate a better quality--size tradeoff. \add{Interactive per-scene plots of PSNR, SSIM, and LPIPS are available on our project website.\footnote{\url{https://yuang-ian.github.io/liftgs/}}}
We have several observations.

\textit{Monolithic and Single fail the streaming requirements.} Monolithic \add{gives the highest PSNR at $L_3$} but transmits an independent full model per tier (up to \SI{+1.9}{\giga\byte} cumulatively on Mip-360), violating \textit{quality scalability}. Single instead renders a single full-resolution model at all levels, but its quality collapses at coarse levels (\SI{17.33}{\decibel} on T\&T's $L_0$) from overfitting.
Neither method is usable in a streaming scenario.

\begin{figure*}
    \centering
    \includegraphics[width=0.96\linewidth]{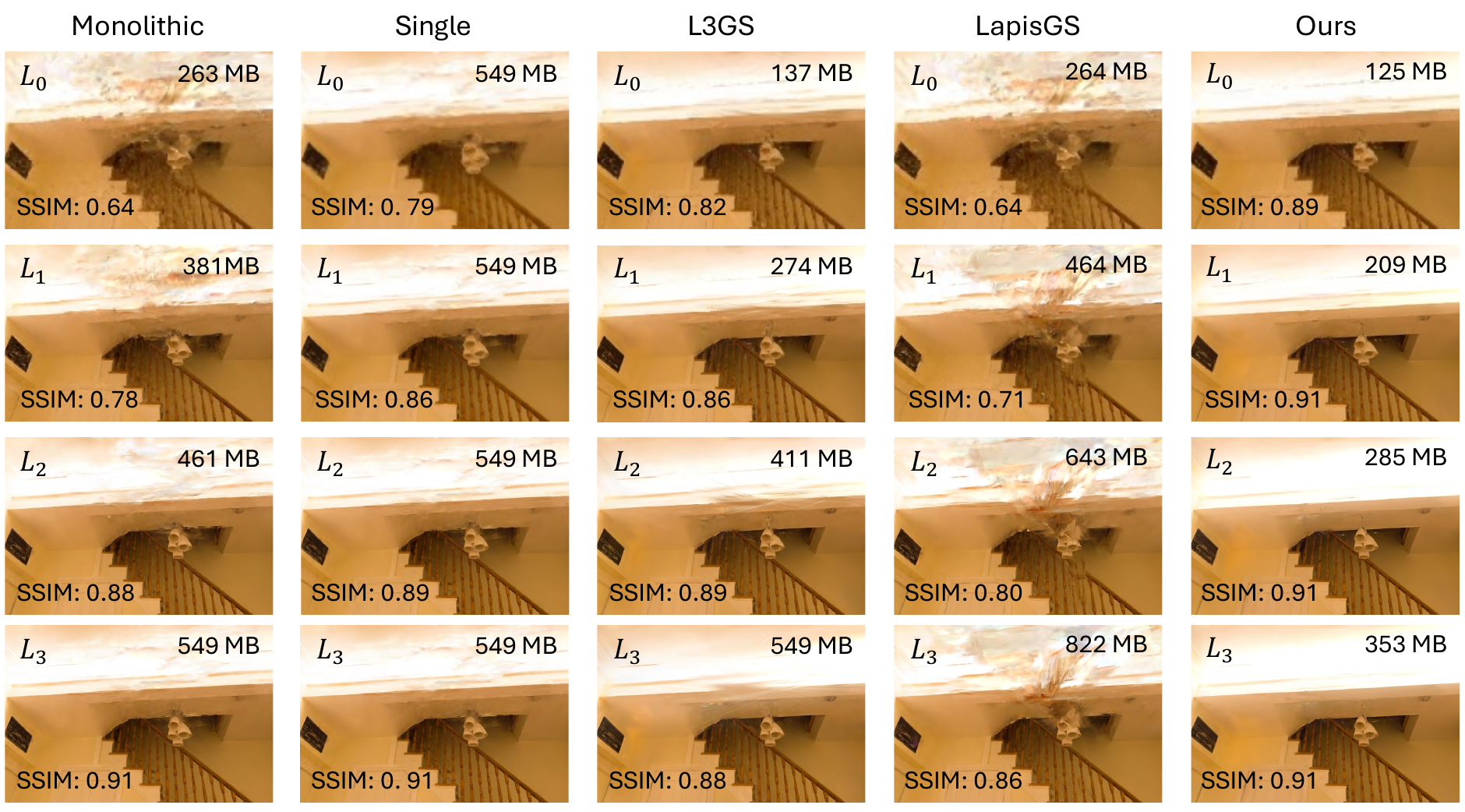}
    \caption{\textbf{Sample renderings of \textit{Playroom}.} LapisGS accumulates the geometry error from low layers. L3GS stalls because frozen ancestors plus a fixed splat budget leave nothing useful to add.
    }
    \label{fig:qualitative_playroom}
\end{figure*}


\textit{{\name} achieves better visual quality than independent layer baselines.} 
The $\Delta$ rows in \cref{tab:quantitative_results_all} show {\name} improves PSNR over the strongest baseline at each level by \SI{0.31}{\decibel} on average, peaking at \SI{+0.69}{\decibel} (T\&T $L_1$); SSIM and LPIPS follow. The second-best position alternates between LapisGS and L3GS, so the gain is structural to the wavelet construction, not to any particular layered design. It stems from correcting error accumulation: both baselines freeze lower-layer splats, locking in geometric errors that propagate upward, whereas {\name}'s parent-child construction lets higher levels reposition and correct ancestral splats via the learned $\boldsymbol{\psi}$ and $\boldsymbol{\alpha}$.

\textit{{\name} dramatically reduces both storage and rendering footprint.} Compared to LapisGS, {\name} requires only $40.5\%$ of the storage payload (DB $L_3$) and $17.9\%$ of the rendering footprint (Blender $L_3$). Compared to L3GS, the reductions are $60.9\%$ and $16.3\%$ at the same operating points. The two baselines exhibit different growth patterns. LapisGS's storage inflates unboundedly as densification adds splats per layer, while L3GS's storage is capped by its pretrained-then-pruned initialization but grows linearly with layer count. Yet {\name} dominates both, because neither growth pattern reflects efficient representation.
Also, as shown in \Cref{fig:scatter_memory,fig:scatter_storage}, the {\name} cluster (red stars) remains highly compact in the top-left, while LapisGS (orange) and L3GS (purple) drift rightward across levels.

The root cause is the splat redundancy inherent to independent layers. Unable to structurally adjust flawed lower layers, both LapisGS and L3GS must over-densify higher layers to mask prior errors, producing persistent ``ghost splats''. {\name}'s wavelet construction eliminates this redundancy by dynamically replacing parents with optimized children that progressively refine the same region, with no need to maintain obsolete ancestors once finer descendants exist. Furthermore, the gap between {\name}'s storage and rendering memory (\eg, \SI{405}{\mega\byte} vs.\ \SI{215}{\mega\byte} on DB $L_3$) provides a runtime advantage absent in either independent-layer baseline, where storage and rendering memory are necessarily equal. We further quantify splat redundancy in \cref{sec:experiment:layer_analysis}.

\textbf{Ablation study.}
\Cref{tab:ablation_sym_asym} compares the symmetric and asymmetric constructions. 
Asym consistently outperforms Sym, the PSNR gap growing from \SI{+0.82}{\decibel} at $L_0$ to \SI{+1.01}{\decibel} at $L_3$. This confirms the limitation in \cref{sec:method:design_space}: the rigid $\boldsymbol{\alpha}=\mathbf{1}$ constraint forces children to deviate equally, over-restrictive for non-symmetric geometry, while the per-attribute $\boldsymbol{\alpha}$ recovers the lost expressivity. Despite five extra scalars per node, Asym even matches or undercuts Sym's storage at $L_0$--$L_1$ (higher expressivity needs fewer splits); the $+0.83\%$--$1.94\%$ overhead at $L_2$--$L_3$ is dwarfed by the \SI{>1}{\decibel} gain. \Cref{sec:experiment:layer_analysis} further shows Sym's rigidity spawns more redundant splats.

\begin{table}[thbp]
    \caption{Ablation comparison across all datasets on average at four quality levels. PSNR is shown, while SSIM and LPIPS reveal the same trend.}
    \label{tab:ablation_sym_asym}
    \centering
    \setlength{\tabcolsep}{4pt}
        \begin{tabular}{c|cc|cc}
            \hline
            \multirow{2}{*}{ Method } & \multicolumn{2}{c|}{$L_0$} & \multicolumn{2}{c}{$L_1$} \\
             & PSNR (dB) & Storage (MB) & PSNR (dB) & Storage (MB)\\
            \hline
            Sym. & 28.84 & 105.95 & 28.64 & 188.81 \\
            Asym. & \textbf{29.66} & \textbf{105.15} & \textbf{29.55} & \textbf{188.72} \\
            \hline
            \hline
            \multirow{2}{*}{ Method } & \multicolumn{2}{c|}{$L_2$} & \multicolumn{2}{c}{$L_3$} \\
             & PSNR (dB) & Storage (MB) & PSNR (dB) & Storage (MB)\\
            \hline
            Sym. & 27.87 & 271.35 & 26.72 & 344.49 \\
            Asym. & \textbf{28.86} & \textbf{276.62} & \textbf{27.73} & \textbf{347.36} \\
            \hline
        \end{tabular}
\end{table}


\textbf{Qualitative results.}
\Cref{fig:qualitative_playroom} illustrates rendered outputs on the \textit{Playroom}.
At $L_0$, LapisGS and Monolithic produce severely degraded geometry (SSIM $0.64$) with geometry collapsing into incoherent distributions, due to insufficient supervision from the $8\times$ downsampled images. In LapisGS, this foundational failure propagates upward as frozen artifacts. 
L3GS avoids this by initializing from a pretrained full-resolution model, but its fixed-size frozen layers offer no mechanism for further refinement.
{\name}'s wavelet decomposition enforces a collinear refinement constraint, allowing each level to correct residual errors. This produces coherent geometry early on (SSIM $0.89$ at $L_0$) and scales smoothly to high-fidelity output at $L_3$.

\subsection{Analysis on Inter-Layer Correlation} \label{sec:experiment:layer_analysis}

\begin{figure*}[t!]
    \centering
    \begin{minipage}{0.75\textwidth}
        \centering
        \includegraphics[width=\linewidth]{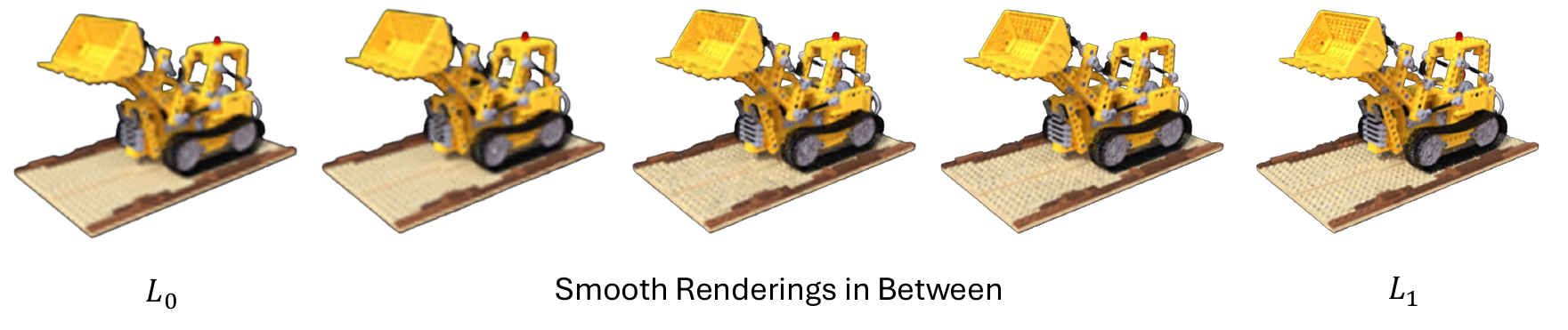}
        \caption{Example of smooth quality transition of \textit{Lego}. The wavelet construction inherently supports smooth quality improvement.}
        \label{fig:smooth_transition}
    \end{minipage}\hfill
    \begin{minipage}{0.22\textwidth}
        \centering
        \includegraphics[width=\linewidth]{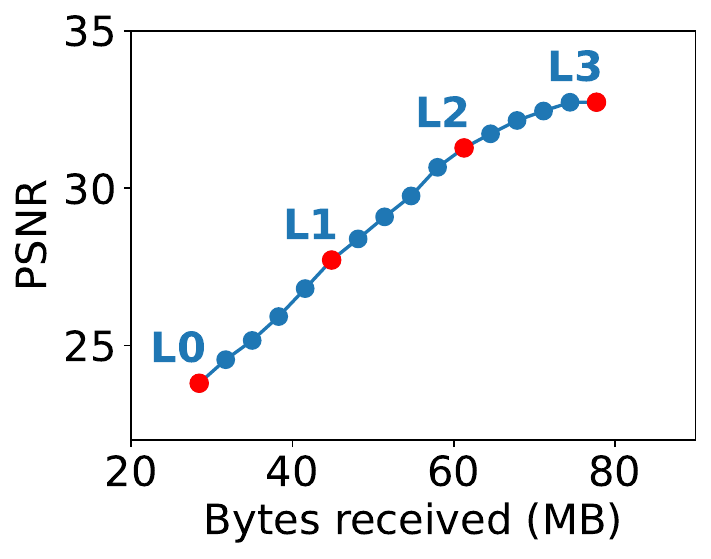}
        \caption{The quality of \textit{Lego} improves smoothly with more bytes received.}
        \label{fig:sweep_psnr_lego_demo}
    \end{minipage}
\end{figure*}

\begin{table}
    \caption{The ratios of transparent splats at four quality levels on average. The opacity threshold is 0.005, same to~\cite{kerbl20233d}.}
    \label{tab:ratio_transparent}
    \centering
    \begin{tabular}{c|ccccc}
        \toprule
               & \multicolumn{4}{c}{\% Transparent Splats} \\
        Method & $L_0$ & $L_1$ & $L_2$ & $L_3$ \\
        \midrule
        LapisGS & 15.92\% & 46.91\% & 55.59\% & 65.83\% \\
        L3GS & 14.95\% & 34.00\% & 39.81\% & 49.73\%\\
        {\name} (Sym.) & 14.40\% & 25.99\% & 33.67\% & 38.85\% \\
        {\name} & \textbf{12.08\%} & \textbf{17.02\%} & \textbf{21.04\%} & \textbf{24.92\%} \\
        \bottomrule
    \end{tabular}
\end{table}

\begin{table}
    \centering
    \caption{Breakdown of the ratios of transparent splats, which shows independent-layer methods fail to efficiently utilize low-layer splats. Ratios larger than 50\% are marked in \textbf{bold}.}
    \label{tab:breakdown_transparent}
    \begin{tabular}{cc|cccc|c}
        \toprule
         & & \multicolumn{4}{c|}{\% Transparent Splats} & \\
         & & Layer 0 & Layer 1 & Layer 2 & Layer 3 & Total \\
        \midrule
        \multirow{4}{*}{\rotatebox{90}{LapisGS}} & $L_0$ & 15.92\% & - & - & - & 15.92\% \\
        & $L_1$ & \textbf{80.33\%} & 16.17\% & - & - & 46.91\% \\
        & $L_2$ & \textbf{90.66\%} & \textbf{78.27\%} & 15.34\% & - & 55.59\% \\
        & $L_3$ & \textbf{93.82\%} & \textbf{88.66\%} & \textbf{72.55\%} & 14.20\% & 65.83\% \\
        \midrule
        \multirow{4}{*}{\rotatebox{90}{L3GS}} & $L_0$ & 14.95\% & - & - & - & 14.95\% \\
        & $L_1$ & \textbf{50.08\%} & 18.93\% & - & - & 34.00\% \\
        & $L_2$ & \textbf{59.76\%} & 43.67\% & 15.99\% & - & 39.81\% \\
        & $L_3$ & \textbf{63.02\%} & \textbf{52.66\%} & \textbf{53.44\%} & 29.82\% & 49.73\% \\
        \bottomrule
    \end{tabular}
\end{table}

\textbf{Splat redundancy.}
To validate the ``ghost splat'' problem, we analyze the ratio of transparent splats (\cref{tab:ratio_transparent}). Independent layers exhibit extreme redundancy: at $L_3$, 65.83\% of LapisGS splats are rendered completely transparent. A layer-by-layer breakdown (\cref{tab:breakdown_transparent}) reveals that 93.82\% of Layer 0 and 88.66\% of Layer 1 splats fade out entirely by $L_3$. Because the optimizer cannot structurally alter frozen ancestral layers, it is forced to drop their opacity to mask geometric errors, artificially inflating the model size and compromising rendering efficiency. L3GS reduces this through its pruning-based control but still reaches 49.73\% transparent at $L_3$, with 63.02\% of its Layer 0 splats faded out, confirming the redundancy is intrinsic to the independent-layer construction rather than to any single method.
{\name} mitigates this by refining rather than overlaying splats, dropping the transparent ratio to 24.92\% at $L_3$. Furthermore, the symmetric variant of {\name} exhibits much higher redundancy (38.85\%) than the asymmetric version, confirming that rigid constraints force the optimizer to spawn and subsequently suppress ill-fitting splats.

\begin{figure}[hbpt]
    \centering
    \begin{subfigure}[b]{0.48\columnwidth}
        \centering
        \includegraphics[width=\textwidth]{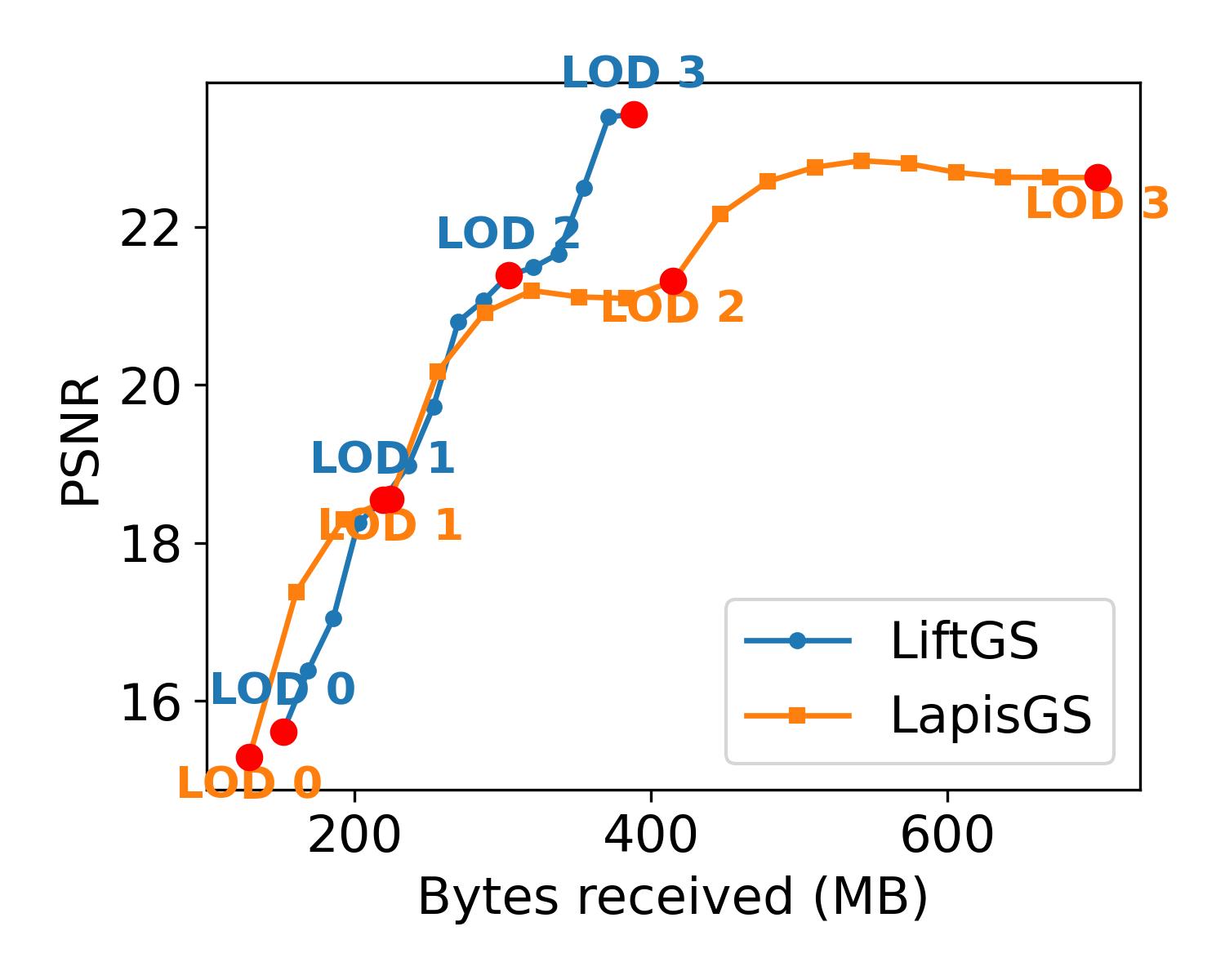}
        \caption{TandT.}
        \label{subfig:tandt_average_PSNR}
    \end{subfigure}
    \begin{subfigure}[b]{0.48\columnwidth}
        \centering
        \includegraphics[width=\textwidth]{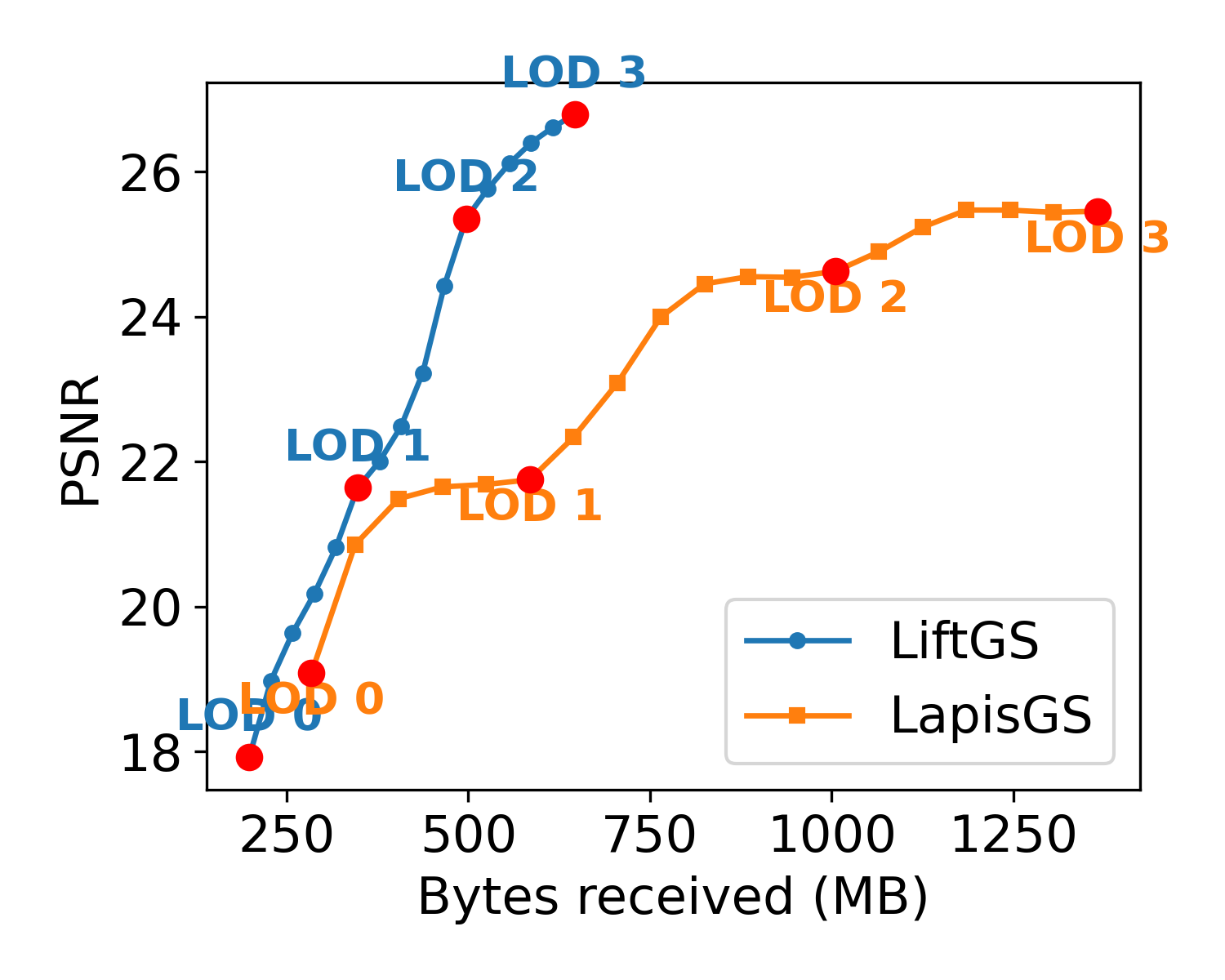}
        \caption{Mip-NeRF360.}
        \label{subfig:360_average_PSNR}
    \end{subfigure}
    \\
    \begin{subfigure}[b]{0.48\columnwidth}
        \centering
        \includegraphics[width=\textwidth]{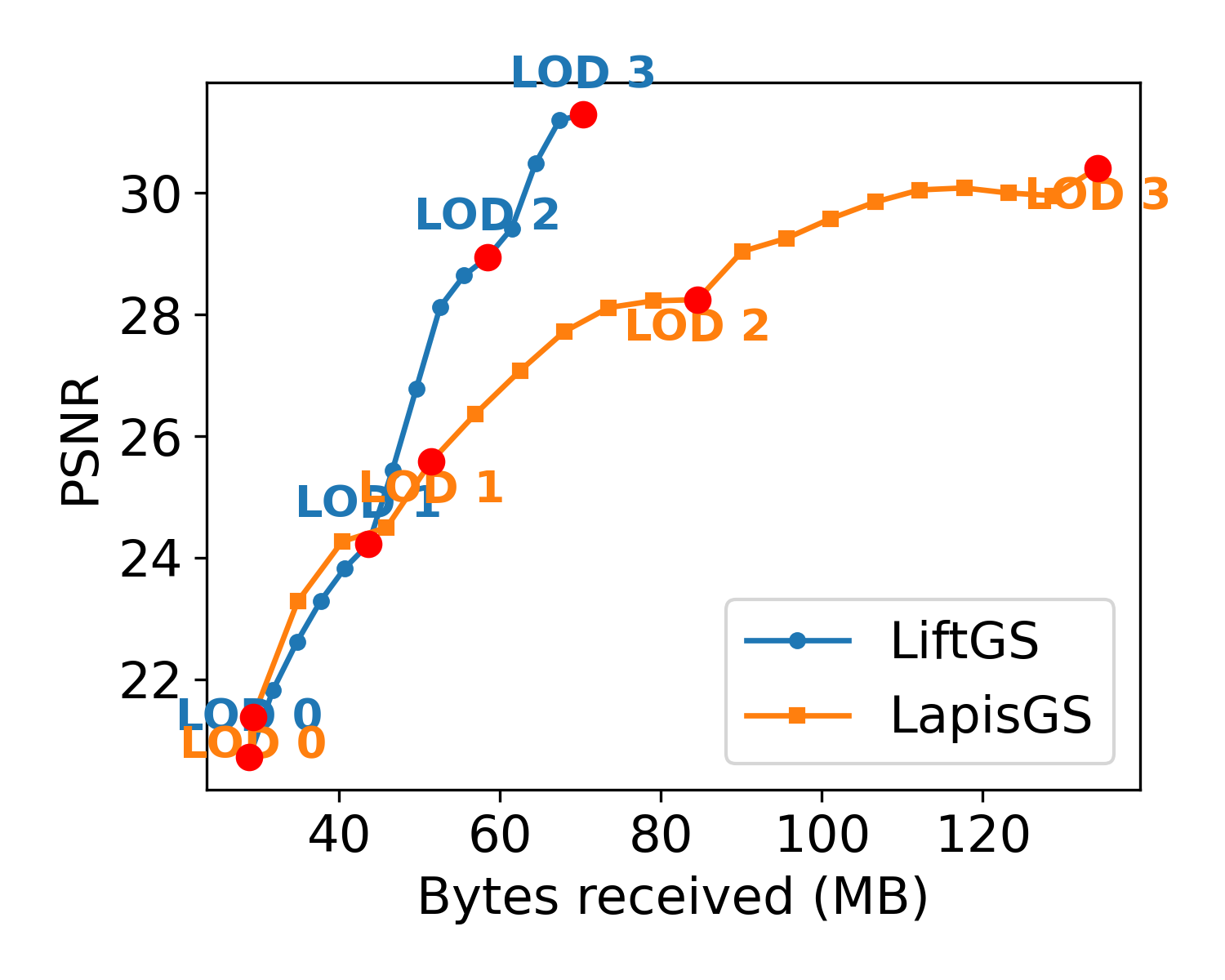}
        \caption{Synthetic Blender.}
        \label{subfig:nerf_synthetic_average_PSNR}
    \end{subfigure}
    \begin{subfigure}[b]{0.48\columnwidth}
        \centering
        \includegraphics[width=\textwidth]{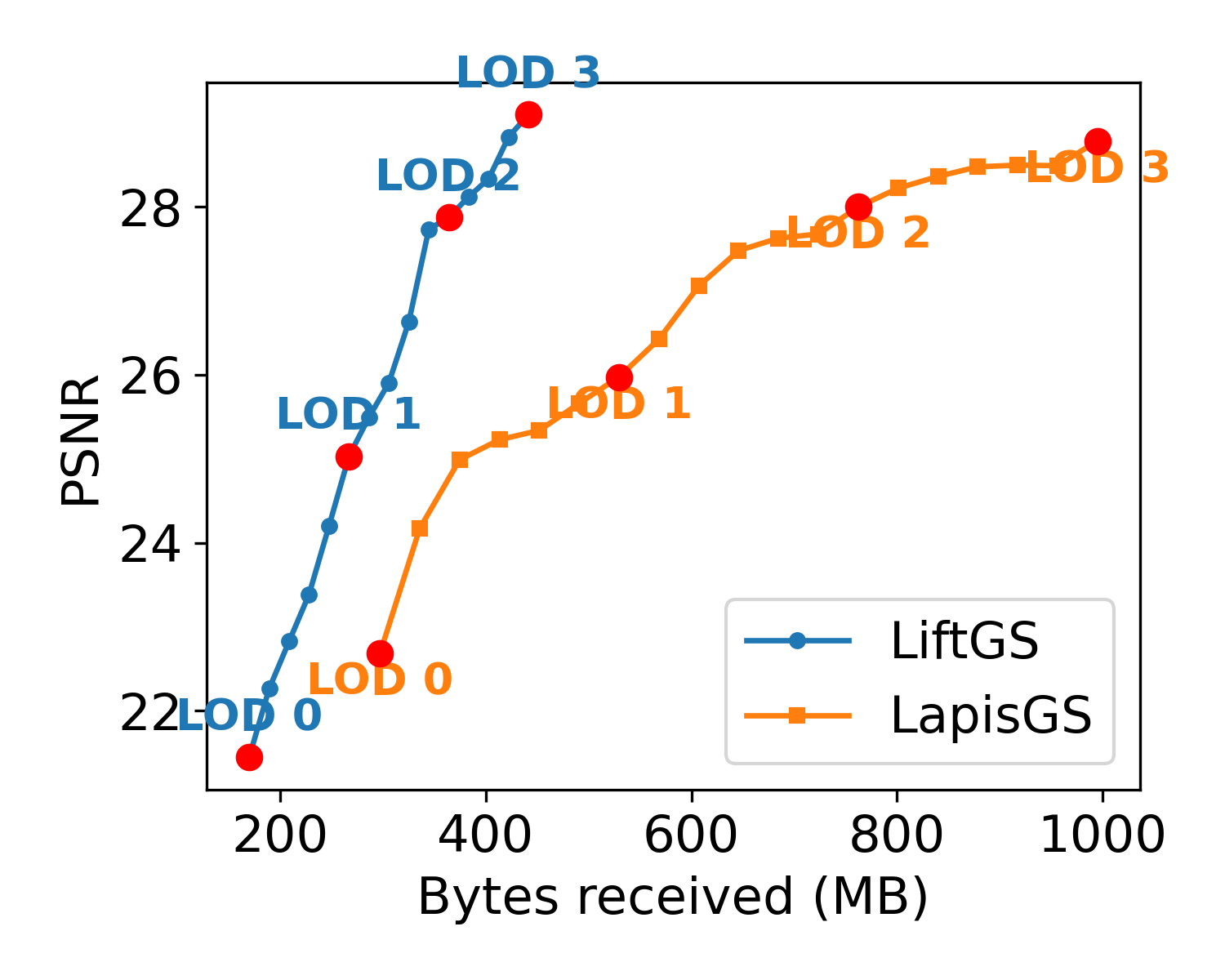}
        \caption{Deep Blending.}
        \label{subfig:db_average_PSNR}
    \end{subfigure}
    \caption{\textbf{Progressive streaming performance of {\name} vs.\ LapisGS}. \add{RD curves at full resolution.} On all datasets, {\name} achieves a smoother quality transition.}
    \label{fig:continuous_transmission}
\end{figure}

\textbf{Quality transition.}
Independent layers overfit to target resolutions, producing step-wise quality jumps (\cref{sec:introduction,sec:method:rendering}). We simulate progressive transmission ($L_0$ to $L_3$) on all datasets over 20 snapshots, ordering {\name}'s splats by refinement energy ($\|\boldsymbol{\psi}\|$) and LapisGS's by opacity\add{, and score every snapshot at full resolution}. The \add{rate-distortion (RD)} curve (\cref{fig:continuous_transmission}) shows {\name} scaling fidelity smoothly with received data, whereas LapisGS stagnates then jumps at level boundaries, wasting its intermediate states. \add{Curves toward the upper left are better.} \Cref{fig:smooth_transition} also shows a visual example and the corresponding \add{RD} curve is presented in \cref{fig:sweep_psnr_lego_demo}.
\add{On Blender and T\&T, the two base layers $L_0$ are of similar size, and LapisGS first sends its most opaque new splats, which raise PSNR quickly. LapisGS is therefore comparable or slightly higher at the lowest rates, until the steeper {\name} curve overtakes it.}

\begin{figure}[t]
    \centering
    \includegraphics[width=0.85\columnwidth]{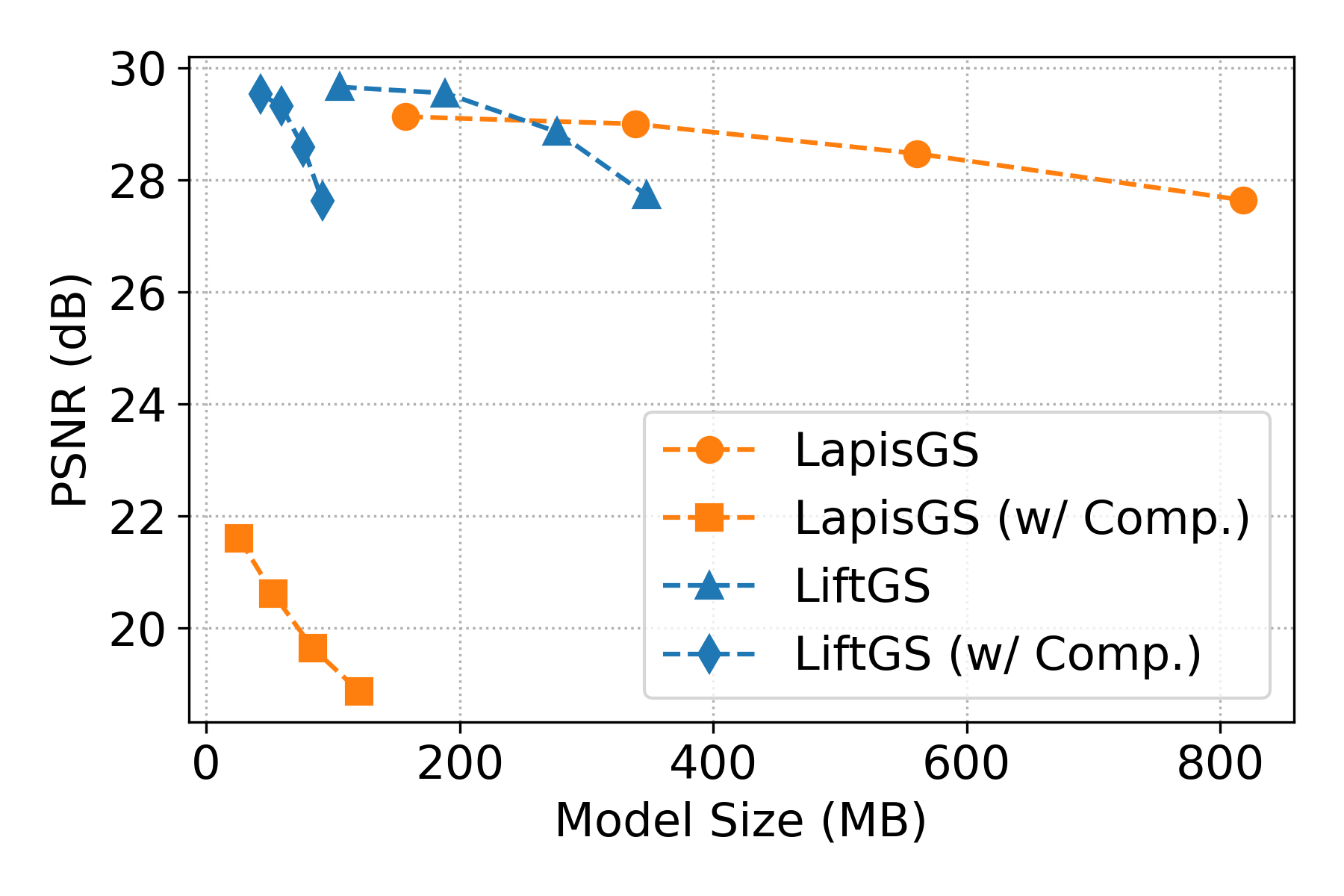}
    \caption{Comparison of compressibility of LapisGS and {\name} (averaged across all datasets). 
    {\name} organizes splats as a wavelet decomposition, which maintains strong sparsity, achieving better compression performance compared with LapisGS under the same compression scheme.}
    \label{fig:compression_proof}
\end{figure}

\textbf{Compressibility.} 
As theorized in \cref{sec:method:design_space} and validated in \cref{fig:energy_concentration}, 
the wavelet detail coefficients $\boldsymbol{\psi}$ exhibit strong sparsity. This structural property naturally yields highly compressible signals. To demonstrate this, we pass {\name} through a basic proof-of-concept compression pipeline: 8-bit uniform scalar quantization followed by zstd entropy coding~\cite{collet2018zstandard}.
The results (\cref{fig:compression_proof}) reveal a dramatic delivery advantage. Uncompressed {\name} already stores less than half of LapisGS at $L_3$ (347.36 MB vs. 818.31 MB). Applying the same pipeline to LapisGS is destructive: because its splats are full, independent attribute sets rather than sparse residuals, 8-bit quantization collapses its PSNR to 18.87 dB while still occupying 120.92 MB. In contrast, the same compression on {\name}'s detail coefficients $\boldsymbol{\psi}$ shrinks the payload to 91.75 MB (11.2\% of uncompressed LapisGS) at a negligible penalty of 0.10 dB. This confirms that the wavelet decomposition not only mitigates structural bloat but actively structures the scene into sparse, compressible detail coefficients ideal for scalable transmission.

\subsection{Client-Side Overhead} \label{sec:experiment:client}

\add{Unlike LapisGS, whose layers are plain splats appended to the render buffer, {\name} must reconstruct the new leaves from $\mathbf{P}$, $\boldsymbol{\psi}$, and $\boldsymbol{\alpha}$ before they can be rendered. We measure this overhead with a client that keeps the tree as flat arrays in GPU memory; the tree is needed only to apply refinements, while the rasterizer reads only the leaf buffer, whose size is the rendering footprint (Mem.). A level transition consists of three steps: uploading the received refinement to the GPU; updating the tree, which inserts the new nodes and writes their $\boldsymbol{\psi}$ and $\boldsymbol{\alpha}$; and constructing the active leaves with \cref{eq:asymmetric} into the buffer that the rasterizer reads. Each step ends with a GPU synchronization, we report the average of five runs, and the constructed leaves match the reference reconstruction exactly on all scenes.}
\add{As shown in \cref{tab:client_cost}, reconstruction costs 4--6\,ms per level transition on average, comparable to the 4--5\,ms needed to upload the refinement to the GPU. It is incurred once per transition rather than per frame, and the reconstructed leaves are rendered by the unmodified 3DGS rasterizer at 5.8--9.7\,ms per frame on average. The overhead of the tree therefore does not prevent real-time updating and rendering.}

\begin{table}[t]
    \caption{\add{Client-side cost of {\name} per level transition, averaged over the four datasets. Reconstruction consists of updating the tree (inserting the new nodes and writing their $\boldsymbol{\psi}$ and $\boldsymbol{\alpha}$) and building the splats (constructing the active leaves read by the rasterizer). Rendering: time per frame at full resolution at the new level.}}
    \label{tab:client_cost}
    \centering
    \add{%
    \begin{tabular}{c|c|cc|c}
        \toprule
        & Upload & \multicolumn{2}{c|}{Reconstruction (ms)} & Rendering \\
        Transition & (ms) & Update Tree & Build Splats & (ms/frame) \\
        \midrule
        $L_0\!\to\!L_1$ & 4.2 & 2.5 & 1.5 & 5.8 \\
        $L_1\!\to\!L_2$ & 4.9 & 2.8 & 2.1 & 7.7 \\
        $L_2\!\to\!L_3$ & 5.9 & 3.6 & 2.3 & 9.7 \\
        \bottomrule
    \end{tabular}}
\end{table}


%% file: sec/5-conclusion.tex
\section{Conclusion and Discussion} \label{sec:conclusion}

We presented {\name}, a multiresolution 3D Gaussian representation designed for progressive delivery and scalable client rendering. By organizing Gaussian splats through a lifting scheme, {\name} represents the scene as a critically-sampled and coarse-to-fine wavelet decomposition in which each finer splat is generated from a coarser one by a learned detail coefficient. This allows one representation to serve multiple quality levels while separating cumulative transmission payload from active-splat memory. Across four datasets, the representation improves the quality--resource tradeoff relative to the evaluated layered baselines, reduces transparent-splat accumulation, and supports smoother quality improvements during simulated progressive delivery. The learned detail coefficients also support substantial additional payload reduction with a simple quantization and entropy-coding pipeline. \add{On the client, applying the refinements of a new level takes a few milliseconds.} These findings establish representation-level benefits for scalable 3DGS streaming\add{. End-to-end evaluation over real networks and bandwidth-aware scheduling are left for future work.}

The refinement tree also opens up several interesting directions for future work. One is region-of-interest streaming: selectively refining branches that contribute to a chosen object or viewing region could concentrate the available bandwidth and rendering memory on content of greater interest, while preserving a coarse representation elsewhere. Another is perceptually guided refinement: prioritizing splits by their expected contribution to visible image quality per transmitted byte could make early updates more useful than ordering them by coefficient magnitude alone. Both directions build on the ability to select and transmit individual refinements within the same trained representation.